\documentclass{article}
\PassOptionsToPackage{numbers,sort&compress}{natbib}
\usepackage[main,final]{neurips_2026}
\usepackage[T1]{fontenc}
\usepackage[utf8]{inputenc}
\usepackage{amsmath,amssymb,bm}
\usepackage{graphicx,booktabs,array}
\usepackage{tikz}
\usepackage{subcaption}
\usepackage{float,wrapfig}
\usepackage{enumitem}
\usepackage{xcolor,microtype}
\usepackage{hyperref,url}
\DeclareRobustCommand{\sname}{FATE}
\DeclareRobustCommand{\lname}{Flow from Adaptive Token Estimation}

\newsavebox{\fatetablebox}
\newcommand{\fitwidth}[1]{\sbox{\fatetablebox}{#1}\ifdim\wd\fatetablebox>\linewidth
    \resizebox{\linewidth}{!}{\usebox{\fatetablebox}}\else\usebox{\fatetablebox}\fi}
\title{Dense Flow from Adaptive Correspondence}
\author{Cheng-Lin Hsieh \qquad Stella X. Yu \\
  University of Michigan \\
  \texttt{\{clhsieh,stellayu\}@umich.edu}
}
\begin{document}
\maketitle

\begin{abstract}
Optical flow requires reliable matching from sufficient visual context
and precise motion at every pixel. Existing methods such as WAFT
match and estimate flow on the same coarse grids, then recover
dense flow through learned upsampling, without explicitly linking
visual support for matching to dense flow reconstruction.
{\it Our key insight is that dense pixel-level flow does not require
equally dense matching.  We decouple correspondence scale from flow
resolution by adapting the size and shape of spatial supports to image
structure for reliable matching, while reusing these supports for
precise dense flow reconstruction. }
We realize this in Flow from Adaptive Token Estimation
(FATE), whose token supports expand within coherent regions and
contract near motion boundaries. Shared pixel-to-token associations
pool features for matching and reconstruct pixel flow from token
motions. Flow supervision jointly trains pixel grouping, token matching, and flow estimation without grouping labels, allowing FATE to anticipate
which pixels are likely to move together from appearance alone.
Trained on FlyingChairs/FlyingThings3D, \sname{} achieves the highest
{\it zero-shot} accuracy on Sintel Final and KITTI, a critical setting given scarce real-world flow labels.  
For visible pixels, \sname{} improves accuracy within coherent regions,
and more near motion boundaries. Its largest gains occur where direct
matches are unavailable: Occluded or out-of-frame pixels account for
roughly 10\% of valid pixels but 50\% of WAFT's total error.
Experiments show that adaptive correspondence through pixel-to-token
grouping enables reliable matching and precise pixel-level flow,
improving generalization to unseen real scenes.
\end{abstract}

\section{Introduction}

Visual correspondence asks which elements in two images refer to the
same scene content; finding these correspondences is called matching.
We study two-image optical flow, which represents image motion as
displacement from each pixel in the first image to {\it its
corresponding location} in the second, even if it is
occluded or outside the image
\citep{horn1981DeterminingOpticalFlow,lucas1981IterativeImageRegistration}.
This close relationship between matching and flow estimation has led
to a common design where both operate on the same spatial grid.

Yet matching and flow estimation have different demands: Reliable
matching needs sufficient visual context, while flow must be estimated
precisely at every pixel. A distinctive corner is easy to match
across images, making its displacement easy to estimate.
A pixel on a white wall, however, could match many white pixels
in the second image, making its displacement hard to pinpoint.
Yet the wall itself may be easier to identify. Similarly, when part
of an object is occluded or leaves the image, its motion may
be inferred from the visible parts. In both cases, matching
coarser structures can help estimate precise pixel flow where
direct matches are unreliable or unavailable.
The challenge is that broader spatial support can resolve ambiguous
matches but risks mixing pixels with distinct motions, while precise
flow must distinguish pixels on opposite sides of motion boundaries
\citep{black1996RobustEstimationMultiple,ren2008LocalGroupingOpticalFlow}.

Modern flow methods address these demands in two stages.
They aggregate information from larger regions for reliable
matching, but matching and flow estimation remain tied to the same
fixed grid, often at reduced resolution for efficiency
\citep{teed2020RAFTRecurrentAllPairsa,xu2022GMFlowLearningOpticala,
huang2022FlowFormerTransformerArchitecturea,wang2025WAFTWarpingAloneField}.
Learned upsampling then recovers precise dense flow from coarse
estimates using pixel-specific weights
\citep{teed2020RAFTRecurrentAllPairsa,wang2025WAFTWarpingAloneField}.
Ideally, the pixels contributing to a match should receive flow
estimates from that match. Yet despite end-to-end training, these
methods do not explicitly link these pixels to the weights used
for dense flow reconstruction.
This raises the question of whether their implicit relationship
generalizes from synthetic training data to unseen real scenes.

{\it Our key insight is that dense pixel-level flow does not require
equally dense matching.  We decouple correspondence scale from flow
resolution by adapting the size and shape of spatial supports to image
structure for reliable matching, while reusing these supports for
precise dense flow reconstruction. }
{\bf 1) Correspondence should adapt} to image structure,
matching distinctive visual entities through spatial supports
that make them reliably identifiable.
These matches provide flow estimates for dense reconstruction.
{\bf 2) The same grouping should govern} which pixels contribute to
each match and how each match contributes to their flow estimates.
This links matching and reconstruction without requiring
pixels in a group to move identically.
{\bf 3) Grouping should anticipate common fate from visual appearance.}
The Gestalt principle of \emph{common fate} holds that elements moving
together tend to group together
\citep{wertheimer1938LawsOrganizationPerceptual,yuille1988ComputationalTheoryPerception}.
Yet matching requires deciding what belongs together {\it before}
knowing how it moves. Prior work has used motion to learn
appearance-based object segmentation
\citep{liu2021EmergenceObjectness,lian2023BootstrappingObjectness}.
We instead use flow supervision to {\it learn appearance-based grouping for
correspondence}, allowing the model to predict groups in unseen
images for more reliable and precise flow estimation.

\begin{figure}[hp]
\centering
\begingroup
\color{black}
\definecolor{cropContext}{HTML}{EDBC32}
\definecolor{cropContextEdge}{HTML}{7A5A0E}
\definecolor{cropDetail}{HTML}{56B4E9}
\definecolor{cropDetailEdge}{HTML}{1D5A85}
\providecolor{panelFrame}{gray}{0.72}
\providecolor{learnedHalo}{HTML}{252525}
\providecolor{latticeCell}{HTML}{242424}
\tikzset{
  crop casing/.style={draw=#1Edge, line width=1.4pt, line join=miter},
  crop core/.style={draw=#1, line width=0.9pt, line join=miter},
  crop casing dashed/.style={dash pattern=on 2.9pt off 0.9pt, dash phase=0.25pt},
  crop core dashed/.style={dash pattern=on 2.4pt off 1.4pt},
  learned lines/.style={draw=learnedHalo, double=white, double distance=0.143pt, line width=0.04pt},
  nominal lines/.style={draw=latticeCell, opacity=.48, line width=0.15pt},
  unit contour/.style={draw=learnedHalo, double=white, double distance=0.598pt, line width=0.14pt, line join=round},
  unit cell/.style={draw=white, double=latticeCell, double distance=0.6pt, line width=0.175pt,
    line join=miter, dash pattern=on 2.22pt off 0.96pt}}
\providecommand{\pixelscope}[3]{\pgfpointdiff{\pgfpointanchor{im}{south west}}{\pgfpointanchor{im}{north east}}\pgfgetlastxy{\cropimagewidth}{\cropimageheight}\pgfmathsetlengthmacro{\croppixelx}{\cropimagewidth/#1}\pgfmathsetlengthmacro{\croppixely}{\cropimageheight/#2}\begin{scope}[shift={(im.north west)}, x=\croppixelx, y=-\croppixely]#3\end{scope}}
\providecommand{\cropinset}[3]{\draw[#3] ([xshift=#2,yshift=#2]#1.south west) rectangle ([xshift=-#2,yshift=-#2]#1.north east);}
\providecommand{\cropframe}[2]{\cropinset{#1}{0.4pt}{draw=#2, line width=0.8pt, line join=miter}}
\providecommand{\panelframe}[1]{\cropinset{#1}{0.2pt}{draw=panelFrame, line width=0.4pt}}
\providecommand{\croplocatorpath}[3]{\draw[#1] ([xshift=0.7pt,yshift=-0.7pt]#2) rectangle ([xshift=-0.7pt,yshift=0.7pt]#3);}
\providecommand{\croplocator}[4][]{\if\relax\detokenize{#1}\relax
    \croplocatorpath{crop casing=#2}{#3}{#4}\croplocatorpath{crop core=#2}{#3}{#4}\else
    \croplocatorpath{crop casing=#2, crop casing dashed}{#3}{#4}\croplocatorpath{crop core=#2, crop core dashed}{#3}{#4}\fi}
\providecommand{\cropbox}[6][]{\croplocator[#1]{#2}{#3,1-#4}{#5,1-#6}}
\providecommand{\cropview}[4][]{\begin{tikzpicture}[baseline=(im.south),remember picture]
    \node[inner sep=0pt,outer sep=0pt,anchor=south west] (im) at (0,0)
      {\includegraphics[#2]{#3}};
    \useasboundingbox (im.south west) rectangle (im.north east);
    \begin{scope}
      \clip (im.south west) rectangle (im.north east);
      \begin{scope}[x={(im.south east)},y={(im.north west)}]#1\end{scope}
    \end{scope}
    \if\relax\detokenize{#4}\relax\panelframe{im}\else\cropframe{im}{#4}\fi
  \end{tikzpicture}}
\providecommand{\cropanchors}[5]{\coordinate (#1-nw) at (#2,1-#3); \coordinate (#1-ne) at (#4,1-#3);
  \coordinate (#1-sw) at (#2,1-#5); \coordinate (#1-se) at (#4,1-#5);}
\providecommand{\croplabel}[2]{\tikz[remember picture,baseline=(#1.base)]\node[inner sep=0pt,outer sep=0pt](#1){\strut #2};}
\providecommand{\cropknock}[1]{([shift={(-0.9pt,-0.3pt)}]#1.south west) rectangle ([shift={(0.9pt,0.3pt)}]#1.north east)}
\providecommand{\cropknockpanel}[1]{(#1-sw) rectangle (#1-ne)}
\tikzset{
  crop connector/.style={draw=#1, line width=0.5pt},
  crop connector cased/.style={crop connector=#1, opacity=.75, preaction={draw=#1Edge, line width=1.1pt, opacity=.45}}}
\providecommand{\cropconnectlines}[7]{\begin{scope}[even odd rule]
    \clip (-50cm,-50cm) rectangle (50cm,50cm) #6 #7;
    \draw[crop connector=#1] (#2) -- (#3) (#4) -- (#5);
  \end{scope}
  \begin{scope}
    \clip #7;
    \draw[crop connector cased=#1] (#2) -- (#3) (#4) -- (#5);
  \end{scope}}
\providecommand{\cropconnect}[7]{\begin{tikzpicture}[remember picture,overlay]\cropconnectlines{#1}{#2}{#3}{#4}{#5}{#6}{#7}\end{tikzpicture}}

\setlength{\tabcolsep}{0pt}
\newlength{\groupviewheight}
\setlength{\groupviewheight}{\dimexpr(\linewidth-7.5pt)/10\relax}
\newcommand{\regionalview}[4][]{\cropview[#1]{height=\groupviewheight,interpolate=false}{figures/introduction/panels/figure2_#2_#3.png}{#4}}
\newcommand{\regionaloverlay}[4][]{\regionalview[\input{figures/introduction/panels/figure2_#2_#3_overlay}#1]{#2}{#3}{#4}}
\newcommand{\selectedunit}[3][]{\cropview[\input{figures/introduction/panels/figure2_#2_#3_overlay}#1]{height=\groupviewheight,interpolate=false}{figures/introduction/panels/figure2_#2_#3.png}{cropDetail}}
{\scriptsize
\begin{tabular}{@{}c@{\hspace{1.5pt}}c@{\hspace{1.5pt}}c@{\hspace{1.5pt}}c@{\hspace{1.5pt}}c@{\hspace{1.5pt}}c@{}}
\strut Image 1 & Learned groups & Token (RGB) & GT flow + grid & GT flow + groups & Token (flow) \\[2pt]
\regionalview[\cropbox{cropContext}{0.2321}{0.3929}{0.5179}{0.6786}\cropanchors{fig2-image1-box}{0.2321}{0.3929}{0.5179}{0.6786}\cropanchors{fig2-image1}{0}{0}{1}{1}]{b}{image1}{} &
\regionalview[\input{figures/introduction/panels/figure2_b_groups_overlay}\cropbox{cropDetail}{0.1719}{0.2500}{0.3281}{0.5625}\cropanchors{fig2-groups-box}{0.1719}{0.2500}{0.3281}{0.5625}\cropanchors{fig2-groups}{0}{0}{1}{1}]{b}{groups}{cropContext} &
\cropview[\pixelscope{20}{20}{\draw[unit contour] (12.5,7)--(13.5,7)--(14.5,7)--(15,7.5)--(15,8.5)--(15,9.5)--(14.5,10)--(14,10.5)--(13.5,11)--(12.5,11)--(12,11.5)--(11.5,12)--(11,12.5)--(10.5,13)--(9.5,13)--(9,13.5)--(8.5,14)--(7.5,14)--(6.5,14)--(6,13.5)--(6,12.5)--(6,11.5)--(6,10.5)--(6.5,10)--(7,9.5)--(7.5,9)--(8.5,9)--(9,8.5)--(9.5,8)--(10.5,8)--(11.5,8)--(12,7.5)--cycle;
\draw[unit cell] (10,10)--(14,10)--(14,14)--(10,14)--cycle;}\cropanchors{fig2-unit}{0}{0}{1}{1}]{height=\groupviewheight,interpolate=false}{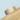}{cropDetail} &
\regionalview[\pixelscope{128}{64}{\draw[nominal lines] (4,0)--(4,64) (8,0)--(8,64) (12,0)--(12,64) (16,0)--(16,64) (20,0)--(20,64) (24,0)--(24,64) (28,0)--(28,64) (32,0)--(32,64) (36,0)--(36,64) (40,0)--(40,64) (44,0)--(44,64) (48,0)--(48,64) (52,0)--(52,64) (56,0)--(56,64) (60,0)--(60,64) (64,0)--(64,64) (68,0)--(68,64) (72,0)--(72,64) (76,0)--(76,64) (80,0)--(80,64) (84,0)--(84,64) (88,0)--(88,64) (92,0)--(92,64) (96,0)--(96,64) (100,0)--(100,64) (104,0)--(104,64) (108,0)--(108,64) (112,0)--(112,64) (116,0)--(116,64) (120,0)--(120,64) (124,0)--(124,64) (0,2)--(128,2) (0,6)--(128,6) (0,10)--(128,10) (0,14)--(128,14) (0,18)--(128,18) (0,22)--(128,22) (0,26)--(128,26) (0,30)--(128,30) (0,34)--(128,34) (0,38)--(128,38) (0,42)--(128,42) (0,46)--(128,46) (0,50)--(128,50) (0,54)--(128,54) (0,58)--(128,58) (0,62)--(128,62);}\cropbox{cropDetail}{0.1719}{0.2500}{0.3281}{0.5625}]{b}{nominal_gt}{cropContext} &
\regionalview[\input{figures/introduction/panels/figure2_b_learned_gt_overlay}\cropbox{cropDetail}{0.1719}{0.2500}{0.3281}{0.5625}]{b}{learned_gt}{cropContext} &
\cropview[\pixelscope{20}{20}{\draw[unit contour] (12.5,7)--(13.5,7)--(14.5,7)--(15,7.5)--(15,8.5)--(15,9.5)--(14.5,10)--(14,10.5)--(13.5,11)--(12.5,11)--(12,11.5)--(11.5,12)--(11,12.5)--(10.5,13)--(9.5,13)--(9,13.5)--(8.5,14)--(7.5,14)--(6.5,14)--(6,13.5)--(6,12.5)--(6,11.5)--(6,10.5)--(6.5,10)--(7,9.5)--(7.5,9)--(8.5,9)--(9,8.5)--(9.5,8)--(10.5,8)--(11.5,8)--(12,7.5)--cycle;
\draw[unit cell] (10,10)--(14,10)--(14,14)--(10,14)--cycle;}]{height=\groupviewheight,interpolate=false}{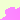}{cropDetail} \\[3pt]
\regionalview[\cropbox{cropContext}{0.3125}{0.4375}{0.6875}{0.8125}]{b3}{image1}{} &
\regionalview[\input{figures/introduction/panels/figure2_b3_groups_overlay}\cropbox{cropDetail}{0.2083}{0.1250}{0.4167}{0.5417}]{b3}{groups}{cropContext} &
\cropview[\pixelscope{20}{20}{\draw[unit contour] (10.5,9)--(11,9.5)--(11,10.5)--(11,11.5)--(11,12.5)--(10.5,13)--(10,12.5)--(10,11.5)--(10,10.5)--(10,9.5)--cycle;
\draw[unit cell] (8,8)--(12,8)--(12,12)--(8,12)--cycle;}]{height=\groupviewheight,interpolate=false}{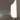}{cropDetail} &
\regionalview[\pixelscope{96}{48}{\draw[nominal lines] (4,0)--(4,48) (8,0)--(8,48) (12,0)--(12,48) (16,0)--(16,48) (20,0)--(20,48) (24,0)--(24,48) (28,0)--(28,48) (32,0)--(32,48) (36,0)--(36,48) (40,0)--(40,48) (44,0)--(44,48) (48,0)--(48,48) (52,0)--(52,48) (56,0)--(56,48) (60,0)--(60,48) (64,0)--(64,48) (68,0)--(68,48) (72,0)--(72,48) (76,0)--(76,48) (80,0)--(80,48) (84,0)--(84,48) (88,0)--(88,48) (92,0)--(92,48) (0,2)--(96,2) (0,6)--(96,6) (0,10)--(96,10) (0,14)--(96,14) (0,18)--(96,18) (0,22)--(96,22) (0,26)--(96,26) (0,30)--(96,30) (0,34)--(96,34) (0,38)--(96,38) (0,42)--(96,42) (0,46)--(96,46);}\cropbox{cropDetail}{0.2083}{0.1250}{0.4167}{0.5417}]{b3}{nominal_gt}{cropContext} &
\regionalview[\input{figures/introduction/panels/figure2_b3_learned_gt_overlay}\cropbox{cropDetail}{0.2083}{0.1250}{0.4167}{0.5417}]{b3}{learned_gt}{cropContext} &
\cropview[\pixelscope{20}{20}{\draw[unit contour] (10.5,9)--(11,9.5)--(11,10.5)--(11,11.5)--(11,12.5)--(10.5,13)--(10,12.5)--(10,11.5)--(10,10.5)--(10,9.5)--cycle;
\draw[unit cell] (8,8)--(12,8)--(12,12)--(8,12)--cycle;}]{height=\groupviewheight,interpolate=false}{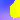}{cropDetail}\cropconnect{cropContext}{fig2-image1-box-ne}{fig2-groups-nw}{fig2-image1-box-se}{fig2-groups-sw}{}{\cropknockpanel{fig2-image1}}\cropconnect{cropDetail}{fig2-groups-box-ne}{fig2-unit-nw}{fig2-groups-box-se}{fig2-unit-sw}{}{\cropknockpanel{fig2-groups}}
\end{tabular}}
\endgroup

\par
\begingroup
\caption{\textbf{\sname{} learns correspondence tokens whose spatial
support adapts to image structure and motion boundaries.}
Token supports spread within coherent regions and sharpen at
motion boundaries, following common motion rather than fixed grid cells.
Learned groups trace slat edges and a bent body outline,
while fixed cells can straddle distinct motions.
The same soft pixel-to-token associations pool features for
matching and combine token motions into dense flow.
White contours show dominant assignments and selected token supports;
dashed squares mark nominal grid cells. Selected supports appear
against RGB images and ground-truth (GT) flow in the token panels.
Gold and blue boxes mark successive enlargements, with matching
RGB and GT flow crops.}

\label{fig:teaser-analysis}
\endgroup
\end{figure}

We realize these ideas in \sname{}, short for \emph{\lname{}}.
{\bf 1) Adaptive correspondence tokens} enable coarse-scale matching
while retaining pixel-level flow prediction.
Each token is anchored to a coarse grid, but its spatial support
adapts in size and shape to image structure. 
As Fig.~\ref{fig:teaser-analysis} shows, 
token supports expand within coherent regions and contract near
motion boundaries, following regions of common motion rather
than fixed grid cells.  Local attention further enriches token
descriptors with broader context, while iterative matching
refines token flow \citep{wang2025WAFTWarpingAloneField}.
{\bf 2) Pixel-to-token grouping} connects matching and dense flow
reconstruction through a shared differentiable association matrix:
Normalized over pixels, it pools features into token descriptors;
normalized over tokens, it combines token flows to reconstruct
precise pixel flow. Each token predicts one motion vector, but each pixel combines motions from multiple tokens, allowing distinct pixel flow estimates within the same token's spatial support.
{\bf 3) Flow supervision teaches appearance-based grouping.}
Pixel-to-token associations, token matching, and dense flow estimation
are learned jointly, without grouping labels. At test time, the
model predicts these associations from appearance alone,
enabling more reliable and precise flow estimation in unseen images.

These ideas deliver stronger zero-shot generalization, a critical need
in optical flow estimation. Dense flow labels are difficult to obtain in real scenes,
so supervised models are often trained on synthetic images and must
transfer to unseen domains without fine-tuning. While visual appearance
may vary across domains, cues to which pixels are likely to move together
may still transfer. Trained on FlyingChairs and FlyingThings3D
\citep{dosovitskiy2015FlowNetLearningOptical,mayer2016LargeDatasetTrain},
\sname{} achieves endpoint errors of 2.10 on Sintel Final and 3.11 on
KITTI, the lowest among published methods in our zero-shot comparison
\citep{butler2012NaturalisticOpenSource,menze2015ObjectSceneFlow}.
Its largest gains occur where direct matches are unavailable:
Occluded or out-of-frame pixels account for only 6.4--13.9\% of valid
pixels, yet contribute 41.9--52.7\% of the  total endpoint error of WAFT\citep{wang2025WAFTWarpingAloneField}.
At equal token counts, \sname{} outperforms fixed-grid alternatives,
benefits more from local attention, and better preserves pixel motion
through pooling and reconstruction of ground-truth flow.
With four refinement steps, it also outperforms five-step WAFT
at lower computational cost.

In summary, we make three contributions.
{\bf 1) Adaptive correspondence tokens} improve flow accuracy
with fewer matching locations through learned spatial supports.
{\bf 2) Shared pixel-to-token associations} provide consistent visual
grounding for matching and dense flow reconstruction.
{\bf 3) Strong zero-shot generalization} is demonstrated across
benchmarks and supported by controlled analyses.
These advances simultaneously enable reliable matching and precise
pixel-level motion estimation, while generalizing effectively
to unseen real-world scenes.
Project website: \href{https://chenglinh.github.io/FATE/}{\sname{}}.

\section{Related work}
\label{sec:related}

\paragraph{Spatial grouping in classical optical flow.}
Classical optical flow uses spatial support to infer motion from ambiguous
local measurements. Lucas--Kanade~\citep{lucas1981IterativeImageRegistration}
pools motion constraints within neighborhoods, while
Horn--Schunck~\citep{horn1981DeterminingOpticalFlow}
couples neighboring estimates through smoothness.
Robust estimation~\citep{black1996RobustEstimationMultiple}
accommodates competing motions.
Ren~\citep{ren2008LocalGroupingOpticalFlow} uses contour-based soft
affinities and figure-ground organization to define spatially adaptive
support for local affine motion estimation near boundaries.
\sname{} learns spatial support jointly with matching features from
flow supervision, using the resulting groups for correspondence and
dense flow estimation.

\paragraph{Common fate and motion-guided grouping.}
The Gestalt principle of common fate holds that elements moving
together are perceived as a group
\citep{wertheimer1938LawsOrganizationPerceptual,yuille1988ComputationalTheoryPerception}.
AMD~\citep{liu2021EmergenceObjectness} jointly learns appearance-based
segmentation and segment motion through video reconstruction,
pooling motion features to predict segment flow.
RCF~\citep{lian2023BootstrappingObjectness} relaxes common fate by
fitting flow with segment motion and within-segment residuals,
then refines grouping from appearance.
These methods learn object segmentation from motion, whereas
\sname{} learns grouping for correspondence and dense flow.
Its soft associations need not follow object boundaries or impose
uniform flow within a token.

\paragraph{Correspondence in deep flow networks.}
Modern two-frame networks largely match features on regular grids.
Learned pyramids and cost volumes~\citep{sun2018PWCNetCNNsOpticala,teed2020RAFTRecurrentAllPairsa},
global matching and cost-memory encoding~\citep{xu2022GMFlowLearningOpticala,huang2022FlowFormerTransformerArchitecturea},
and pretrained representations~\citep{dong2023RethinkingOpticalFlow,zhou2024SAMFlowEliminatingAny}
strengthen features for matching.
Recurrent estimation and learned motion priors~\citep{wang2025SEARAFTSimpleEfficient,luo2024FlowDiffuserAdvancingOptical}
improve flow estimation.  Source-image affinities aggregate motion features~\citep{jiang2021LearningEstimateHidden}, while feature similarity
guides flow propagation~\citep{xu2022GMFlowLearningOpticala}.
Multiscale processing and learned upsampling recover spatial detail~\citep{morimitsu2025DPFlowAdaptiveOptical,jung2023AnyFlowArbitraryScale,
teed2020RAFTRecurrentAllPairsa,wang2025WAFTWarpingAloneField}.
In these methods, pixel support for matching and the weights that
distribute motion to pixels are linked only implicitly.
Building on iterative matching~\citep{wang2025WAFTWarpingAloneField}, \sname{} learns adaptive
correspondence tokens whose shared pixel-to-token associations pool features
for matching and combine token motions into dense flow.

\paragraph{Superpixels and differentiable grouping.}
Superpixel methods partition images into compact regions following
appearance boundaries~\citep{achanta2012SLICSuperpixelsCompared,vandenbergh2012SEEDSSuperpixelsExtracted}.
Learned variants make grouping differentiable and adaptable to
downstream tasks~\citep{jampani2018SuperpixelSamplingNetworks,yang2020SuperpixelSegmentationFully}.
Unlike image segmentation, grouping in \sname{} serves correspondence
and dense motion estimation. Flow supervision jointly learns the
associations and matching features, without requiring pixel-group labels.

\section{\sname{}: \lname{}}
\label{sec:method}

\begin{figure}[t]
\centering
\includegraphics[width=0.99\linewidth]{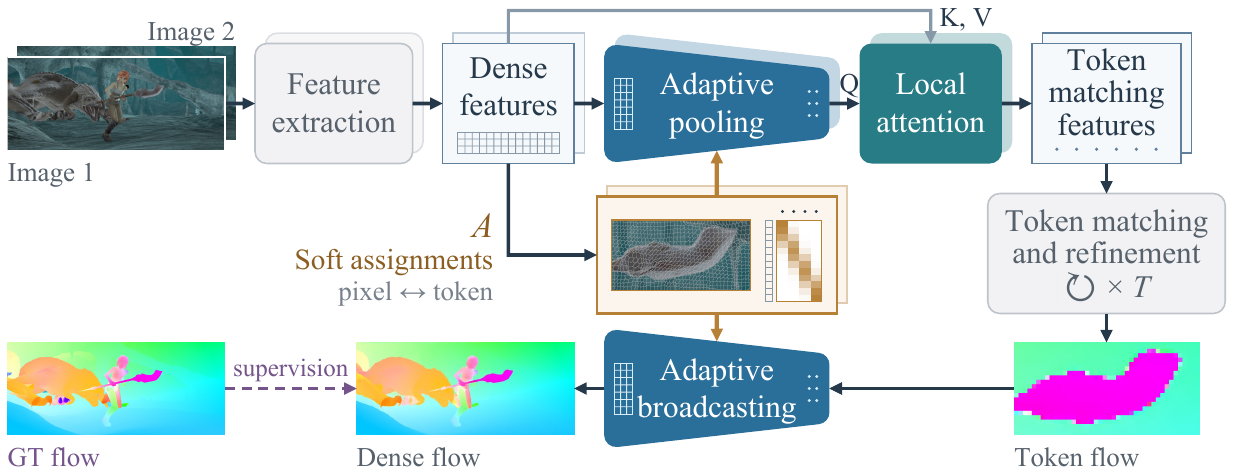}
\caption{\textbf{Shared pixel-to-token associations connect token matching
and dense flow estimation.}
Learned associations $A$ pool dense image features into adaptive
correspondence tokens (center), enriched with nearby context through
local attention before iterative matching across images (right).
The same associations combine estimated token motions into
dense pixel flow (bottom), allowing each pixel to draw on multiple tokens.
Image~2 (lighter branch) has its own associations but shares network
weights with Image~1. Dashed arrows indicate ground-truth flow supervision (Fig.~\ref{fig:unit-spaces}).}
\label{fig:pipeline}
\end{figure}

\paragraph{Overview.}
Given two images, \sname{} estimates dense optical flow by matching
adaptive correspondence tokens at a coarser resolution, then combining
their motions into pixel-level flow (Fig.~\ref{fig:pipeline}).
The model has three components.
{\bf 1) Adaptive correspondence tokens} are anchored to points on a
coarse grid, but their spatial supports are learned from image features.
{\bf 2) Shared pixel-to-token associations} pool each image's dense
features into token descriptors, enriched by local attention for
broader matching context. Iterative matching estimates token motion,
which the same associations combine into pixel flow in the first image.
Since each pixel combines motions from several tokens, pixels within
a token's support need not share identical flow.
{\bf 3) Flow-supervised grouping} jointly learns associations, matching,
and flow estimation from ground-truth pixel flow, without pixel grouping labels.
At test time, the model anticipates which pixels are likely to move
together from appearance alone, enabling reliable matching and precise
pixel flow in unseen scenes.

\paragraph{Adaptive correspondence tokens.}
For an image with $N$ pixels indexed by $x$ and $M$ tokens indexed by $q$,
we learn a pixel-to-token association matrix
$A\in\mathbb R^{N\times M}$ of soft assignments, where $A_{xq}\geq 0$ measures how strongly
pixel $x$ is associated with token $q$.
Each row describes a pixel's associations with tokens, while each
column describes a token's spatial support.

\begin{figure}[t]
\centering
\input{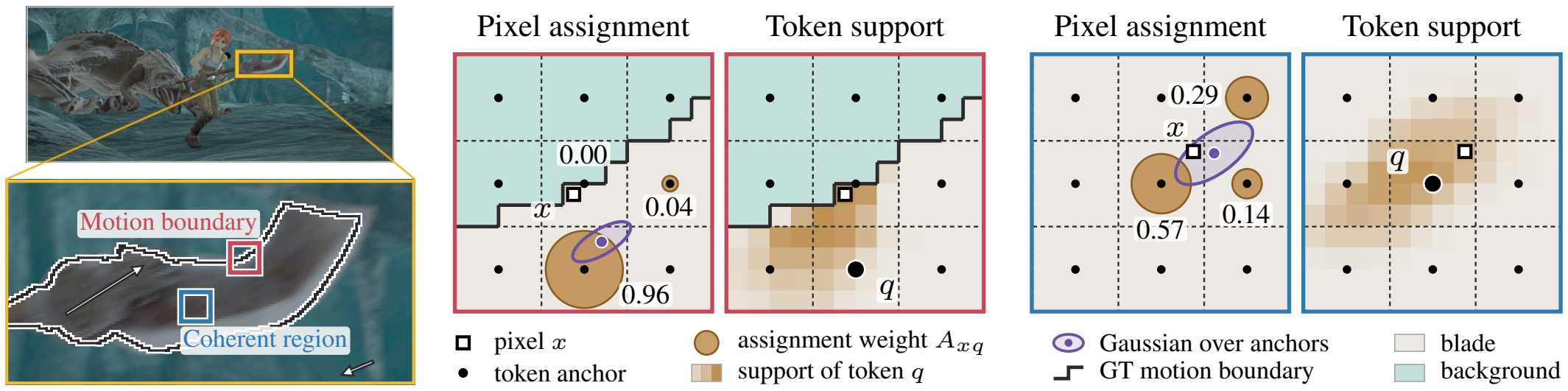}
\caption{\textbf{Learned pixel-to-token associations adapt token support to motion structure.}
\setlength{\fboxsep}{1pt}
Left: the original image and an enlarged view of the moving blade,
with two marked regions: a \fcolorbox{red}{white}{motion boundary}
and a \fcolorbox{blue}{white}{coherent region}.
Arrows indicate ground-truth motion.
Center and right: Each region is shown from two perspectives.
Pixel assignment shows how pixel $x$ associates with nearby tokens
(a row of $A$); token support shows the spatial support of token $q$
(a column of $A$).
Dashed grid lines mark the coarse token lattice, black dots mark
token anchors, and squares mark pixels. Brown disks represent
individual assignment weights $A_{xq}$, with size indicating weight;
purple ellipses show predicted Gaussians, and brown shading shows
token support.
At the motion boundary, $x$ favors a blade token across grid-cell
borders, allowing its support to align with the boundary.
Within the coherent region, $x$ distributes its associations
across several nearby tokens.}
\label{fig:structure}
\end{figure}

\textit{Predicting associations from appearance.}\enskip
Each pixel associates with the $3\times3$ token anchors around
its grid cell. Rather than predicting nine independent scores,
a compact head predicts a two-dimensional Gaussian over these
anchors from dense image features. Its center determines which
nearby tokens the pixel favors, while its shape controls how
widely the associations spread. Evaluating the Gaussian at the
candidate anchors and normalizing over them gives the corresponding
entries of $A$ (Appendix~\ref{app:method-details}).
As Fig.~\ref{fig:structure} illustrates, the resulting token supports
can follow motion boundaries rather than fixed cell borders,
while spreading more broadly within coherent regions.

\begin{figure}[hp]
\centering
\input{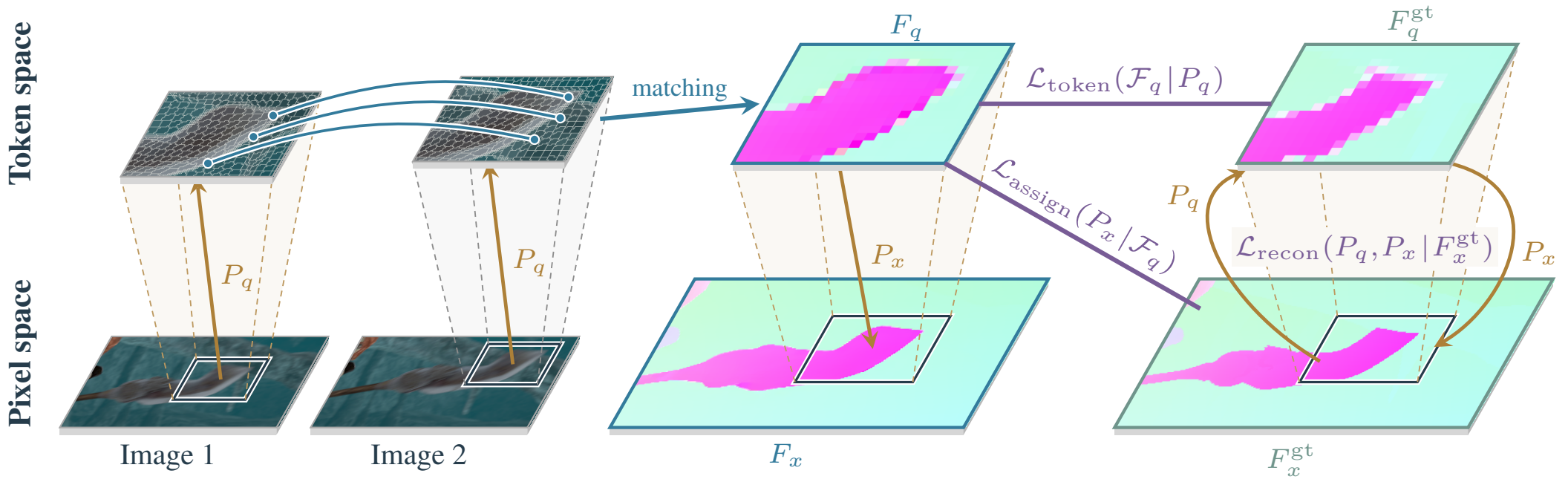}

\caption{\textbf{Matching in token space estimates flow in pixel space
through shared associations and flow supervision.}
Matching operates in coarse token space (top), while dense flow is
formed in pixel space (bottom), across both the image features
(left) and flow predictions (right).
Features from Image~1 and Image~2 are pooled into adaptive tokens
and matched across images (blue links).
On the right, predicted token flow $F_q$ is combined into pixel flow
$F_x$, while ground-truth pixel flow $F_x^{\mathrm{gt}}$ is pooled
into pseudo-ground-truth token flow $F_q^{\mathrm{gt}}$.
The probabilistic predictions $\mathcal F_q$ include token flow
$F_q$ and its uncertainty.
We learn token matching and pixel-to-token associations using only
ground-truth pixel flow, through three complementary losses:
$\mathcal L_{\mathrm{token}}(\mathcal F_q\mid P_q)$ trains token-flow
prediction against pooled ground truth, holding $P_q$ fixed in
constructing the pseudo-ground truth;
$\mathcal L_{\mathrm{assign}}(P_x\mid\mathcal F_q)$ trains associations
to favor tokens that confidently explain pixel motion, holding
token-flow predictions fixed;
and $\mathcal L_{\mathrm{recon}}(P_q,P_x\mid F_x^{\mathrm{gt}})$
trains flow-preserving grouping by pooling and reconstructing fixed
ground-truth flow, independently of matching.
Upper views enlarge the boxed regions, token colors show pooled RGB,
and flow maps share one color scale.}

\label{fig:unit-spaces}
\end{figure}

\paragraph{Shared pixel-to-token associations.}
The same learned associations govern how pixels support their token
matching and how token motions are combined into dense pixel flow
(Fig.~\ref{fig:unit-spaces}).

\textit{1)~Pooling and broadcasting.}\enskip
The association matrix $A$ connects pixel and token spaces in two ways:
Normalized over pixels (columns of $A$), it pools features into token
descriptors; normalized over tokens (rows of $A$), it combines token
motions into pixel flow.
Let $D(B)=\operatorname{diag}(B\mathbf 1)$ denote the diagonal degree matrix
of row sums. Omitting the dependence on $A$, we have 
 two operators:
\begin{align}
\text{pixel-to-token feature pooling:}\quad
&P_q=D(A^\top)^{-1}A^\top,
\label{eq:pool}\\
\text{token-to-pixel flow broadcasting:}\quad
&P_x=D(A)^{-1}A.
\label{eq:broadcast}
\end{align}
The shared associations provide consistent visual grounding for
matching and pixel flow.

\textit{2)~Token features with local attention.}\enskip
Reliable matching requires both features pooled from each token's
support and broader context to resolve ambiguity. Local attention
contextualizes each token's pooled features with surrounding dense
features, sharing network weights for both images.

\textit{3)~Iterative flow refinement and broadcasting.}\enskip
Let $F_q$ and $F_x$ denote token and pixel flow from the first image
to the second. Following the iterative matching framework of WAFT
\citep{wang2025WAFTWarpingAloneField}, each refinement step samples
the second image's token features at current match locations
and updates the token flow. Both images' associations are computed once and reused throughout refinement. At step $t$, the first image's broadcasting operator produces dense pixel flow:
\begin{equation}
F_x^{(t)}=P_x F_q^{(t)}.
\label{eq:inference}
\end{equation}
Each token predicts one motion vector, but each pixel combines
motions from multiple tokens, allowing distinct pixel flow estimates
within the same token's spatial support.

\paragraph{Flow-supervised grouping.}
We jointly learn pixel-to-token associations and token-flow prediction
from ground-truth pixel flow, since no annotation specifies which pixels
should be grouped for correspondence (Fig.~\ref{fig:unit-spaces}).
Three complementary losses supervise token matching, pixel-to-token associations, and preservation of ground-truth pixel motion through pooling and reconstruction. Each has the form $\mathcal L(X\mid Y)$, training $X$ with $Y$ given.

\textit{1)~Token-flow prediction.}\enskip
Since matching operates at the coarse token level, we construct
pseudo-ground-truth token flow by pooling ground-truth pixel flow,
provisionally assuming the current associations are correct.
Following SEA-RAFT~\citep{wang2025SEARAFTSimpleEfficient},
the network predicts token flow $F_q$ and its uncertainty.
We denote these probabilistic predictions by $\mathcal F_q$ and define
$\operatorname{NLL}(\mathcal P;Y)=-\sum_i\log p_{\mathcal P_i}(Y_i)$
as the negative log-likelihood of targets $Y$ under predicted
distributions $\mathcal P$:
\begin{align}
F_q^{\mathrm{gt}}&=P_qF_x^{\mathrm{gt}},\\
\mathcal L_{\mathrm{token}}(\mathcal F_q\mid P_q)
&=\operatorname{NLL}(\mathcal F_q;F_q^{\mathrm{gt}}).
\label{eq:unitloss}
\end{align}
This loss primarily trains token matching, updating token features,
local attention, iterative flow refinement, and uncertainty prediction.
It also trains pixel-to-token associations through gradients from
feature pooling. However, gradients are stopped when constructing
pseudo-ground-truth token flow, preventing the associations from
shifting this supervision to reduce the loss.

\textit{2)~Pixel-to-token associations.}\enskip
Near motion boundaries, neighboring tokens may predict different flows.
Averaging them with spread-out weights in Eq.~\eqref{eq:inference} can produce a
spurious intermediate flow.
To preserve distinct motions, we learn pixel-to-token associations
based on how well each token's predicted flow explains the pixel's
ground truth, accounting for prediction uncertainty.
Provisionally holding token-flow predictions and uncertainties fixed,
we evaluate the likelihood of each pixel's ground-truth flow under
each neighboring token's predicted distribution. The association
weights $P_x$ combine these likelihoods into a mixture, denoted by
$\operatorname{Mix}_{P_x}[\mathcal F_q]$. We minimize
\begin{equation}
\mathcal L_{\mathrm{assign}}(P_x\mid\mathcal F_q)
=\operatorname{NLL}\!\left(
\operatorname{Mix}_{P_x}[\mathcal F_q];F_x^{\mathrm{gt}}
\right).
\label{eq:mixture}
\end{equation}
Unlike averaging token flows, this mixture retains distinct
predictions and uncertainties, encouraging pixels to associate
with tokens that confidently explain their motion.
With token predictions fixed, this loss updates the assignment
head and its input features, not token matching or refinement.

\textit{3)~Flow-preserving reconstruction.}\enskip
Ground-truth pixel flow should also be preserved
when pooled into coarse token flow and reconstructed back at pixel
resolution. We therefore minimize the reconstruction error, with an 
aggregated robust flow penalty $\mathcal D$:
\begin{equation}
\mathcal L_{\mathrm{recon}}(P_q,P_x\mid F_x^{\mathrm{gt}})
=\mathcal D\!\left(
P_xP_qF_x^{\mathrm{gt}},F_x^{\mathrm{gt}}
\right).
\label{eq:flowloss}
\end{equation}
As in superpixel learning~\citep{jampani2018SuperpixelSamplingNetworks},
this loss updates the associations through both $P_q$ and $P_x$,
encouraging the shared pixel-to-token associations to preserve distinct pixel motions.  Since matching is bypassed, the loss trains grouping independently
of token-flow prediction.

\textit{4)~Joint training.}\enskip
We optimize all three losses jointly, applying token-flow and
association losses at each refinement step and reconstruction
loss once (Appendix~\ref{app:method-details}).
Together, they learn token flow and pixel-to-token associations
to achieve both reliable matching and precise pixel-level flow.

\section{Experiments}
\label{sec:experiments}

Our experiments test whether consistent visual grounding improves dense flow
estimation and its transfer to unseen domains. We first
compare zero-shot accuracy with published methods. We then examine where errors
arise and where \sname{} reduces them, focusing on two regions: occluded or
out-of-frame pixels, which lack a direct match and depend on shared evidence,
and pixels near motion boundaries, where distinct motions must be preserved.
A comparison with a fixed-grid model at the same token count isolates the
role of adaptive grouping. Ablations test design choices, and computation profiles
show what fewer tokens make affordable.

\paragraph{Experimental setup.}
All compared methods are trained on
FlyingChairs~\citep{dosovitskiy2015FlowNetLearningOptical} followed by
FlyingThings3D (Things)~\citep{mayer2016LargeDatasetTrain} and evaluated on two
frames.

\textit{1)~Models and training.}\enskip We build \sname{} on WAFT's iterative matching and refinement
framework~\citep{wang2025WAFTWarpingAloneField} with a DINOv3 ViT-S
backbone~\citep{simeoni2025DINOv3}. For an $H\times W$ image, both extract dense
features on an $(H/2)\times(W/2)$ grid, abbreviated $H/2$.
Released WAFT estimates flow on this grid, whereas \sname{} estimates it for
adaptive tokens anchored on an $H/4$ lattice, with a quarter as many tokens.
\sname{} and WAFT use five refinement steps in training and inference unless
stated otherwise (Appendix~\ref{app:compute}).

\textit{2)~Evaluation protocol.}\enskip Things test results are in distribution, while
results on the Sintel~\citep{butler2012NaturalisticOpenSource} and KITTI
2015~\citep{menze2015ObjectSceneFlow} training splits are zero-shot, without
target-domain flow training, as in real scenes where dense flow labels are
difficult to acquire. Things and Sintel each have Clean and Final versions.

\textit{3)~Metrics.}\enskip We report endpoint error (EPE), the Euclidean flow error in pixels averaged per
image and then over images, and KITTI Fl, the percentage of valid pixels whose
error exceeds both 3 pixels and 5\% of the ground-truth motion. Regional
analyses pool endpoint errors over pixels (Appendix~\ref{app:visibility}).
Computation is given in GMACs, billions of multiply--accumulates, per
$540\times960$ image pair.

\paragraph{Zero-shot transfer and regional accuracy.}
We compare overall accuracy with published results for prior methods and
regional accuracy with the released WAFT checkpoint.

\textit{1)~\sname{} achieves the lowest zero-shot EPE on Sintel Final and KITTI.}\enskip In Table~\ref{tab:zero-shot}, \sname{} has lower Sintel Final and KITTI EPE than
every compared method and improves on WAFT in every accuracy column. With a
Depth Anything V2
backbone~\citep{yang2024DepthAnythingV2} instead of DINOv3, \sname{} again has
lower EPE than WAFT on both Sintel and KITTI (Appendix~\ref{app:backbone}).

\begin{figure}[t]
\centering
\input{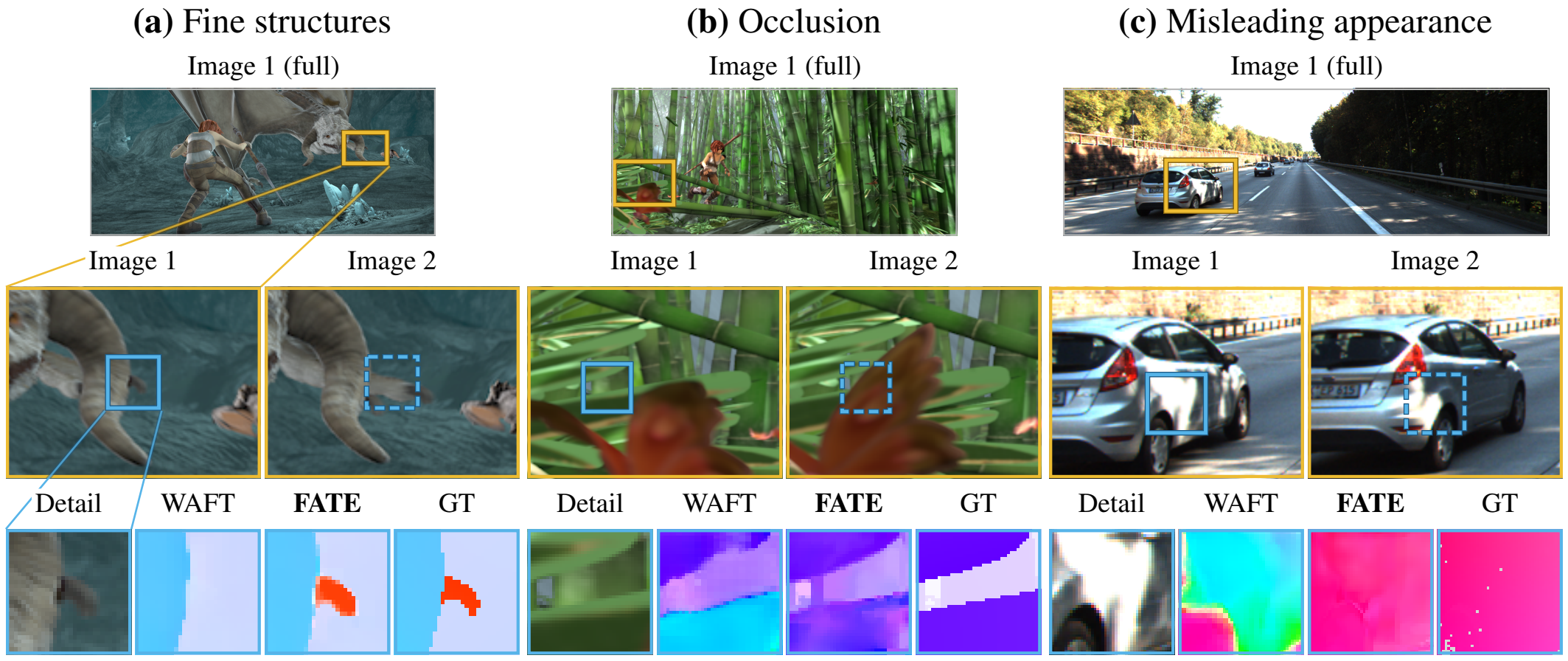}
\begingroup
\caption{\textbf{\sname{} better handles fine structures, occlusion and misleading
appearance.}
(a) A thin protrusion moves unlike the surfaces on both sides; per-pixel
assignments keep its motion distinct.
(b) Leaves hidden by a wing in Image~2 have no match; their tokens infer motion
from surrounding context.
(c) Sunlight and shadows on a car change between frames; WAFT's flow follows
them, while \sname{}'s tokens keep the whole surface's motion coherent.
Gold and blue boxes locate context and detail crops; dashed boxes repeat Image~1
coordinates in Image~2. All predictions are zero-shot; colors share a scale per
example, and gray marks unlabeled GT.}
\label{fig:regions}
\endgroup
\end{figure}

\textit{2)~Pixels without direct matches or near motion boundaries carry most of WAFT's error.}\enskip In Table~\ref{tab:visibility}, occluded or out-of-frame pixels make up only
6.4--13.9\% of the pixels yet account for 41.9--52.7\% of WAFT's summed
endpoint error. Near-boundary pixels likewise account for 10.7--22.9\% of the
error from 3.6--8.7\% of the pixels. Coherent pixels, the remaining
77.8--87.9\%, account for only 35.1--46.7\% of the error, so overall EPE
reveals little about accuracy in the other two regions.
Fig.~\ref{fig:regions} illustrates three cases that make flow hard: a thin
structure near motion boundaries, an occluded background and misleading
lighting. Fig.~\ref{fig:qualitative} shows more such cases over wider
views.

\begin{table}[!tp]
\centering
\footnotesize
\caption{\textbf{\sname{} achieves the lowest zero-shot Sintel Final and KITTI EPE among the compared methods.} All methods train on FlyingChairs and FlyingThings3D. Errors are EPE except KITTI Fl. Bold: lowest error in each accuracy column; --: not reported.}
\label{tab:zero-shot}
\setlength{\tabcolsep}{12pt}
\renewcommand{\arraystretch}{1}
\newcolumntype{N}[1]{>{\hbox to #1\bgroup\hfil}c<{\egroup}}
\begin{tabular}{lN{1.75em}N{1.75em}N{2.25em}N{1.75em}N{2.5em}}
\toprule
Method & \multicolumn{1}{c}{Sintel Clean} & \multicolumn{1}{c}{Sintel Final}
& \multicolumn{1}{c}{KITTI EPE} & \multicolumn{1}{c}{KITTI Fl (\%)}
& \multicolumn{1}{c}{GMACs} \\
\midrule
FlowNet2~\citep{ilg2017FlowNet20Evolution} & 2.02 & 3.14 & 10.10 & 30.4 & 230 \\
PWC-Net~\citep{sun2018PWCNetCNNsOpticala} & 2.55 & 3.93 & 10.40 & 33.7 & 101 \\
RAFT~\citep{teed2020RAFTRecurrentAllPairsa} & 1.43 & 2.71 & 5.04 & 17.4 & 938 \\
GMA~\citep{jiang2021LearningEstimateHidden} & 1.30 & 2.74 & 4.69 & 17.1 & 1352 \\
SKFlow~\citep{sun2022SKFlowLearningOpticala} & 1.22 & 2.46 & 4.27 & 15.5 & 751 \\
DIP~\citep{zheng2022DIPDeepInverse} & 1.30 & 2.82 & 4.29 & 13.7 & 3068 \\
CRAFT~\citep{sui2022CRAFTCrossAttentionalFlow} & 1.27 & 2.79 & 4.88 & 17.5 & 2274 \\
GMFlowNet~\citep{zhao2022GlobalMatchingOverlapping} & 1.14 & 2.71 & 4.24 & 15.4 & 1094 \\
FlowFormer~\citep{huang2022FlowFormerTransformerArchitecturea} & 1.01 & 2.40 & 4.09 & 14.7 & 1715 \\
EMD-L~\citep{deng2023ExplicitMotionDisentangling} & 0.88 & 2.55 & 4.12 & 13.5 & 1755 \\
FlowFormer++~\citep{shi2023FlowFormerMaskedCost} & 0.90 & 2.30 & 3.93 & 14.1 & 1713 \\
MatchFlow(G)~\citep{dong2023RethinkingOpticalFlow} & 1.03 & 2.45 & 4.08 & 15.6 & 1669 \\
AnyFlow~\citep{jung2023AnyFlowArbitraryScale} & 1.10 & 2.52 & 3.76 & 12.4 & \multicolumn{1}{c}{--} \\
RPKNet~\citep{morimitsu2024RecurrentPartialKernel} & 1.12 & 2.45 & \multicolumn{1}{c}{--} & 13.0 & 137 \\
CCMR+~\citep{jahedi2024CCMRHighResolution} & 0.98 & 2.36 & \multicolumn{1}{c}{--} & 12.9 & 12653 \\
SEA-RAFT(L)~\citep{wang2025SEARAFTSimpleEfficient} & 1.19 & 4.11 & 3.62 & 12.9 & 655 \\
SAMFlow~\citep{zhou2024SAMFlowEliminatingAny} & 0.87 & 2.11 & 3.44 & 12.3 & 9717 \\
FlowDiffuser~\citep{luo2024FlowDiffuserAdvancingOptical} & \textbf{0.86} & 2.19 & 3.61 & 11.8 & 2466 \\
DPFlow~\citep{morimitsu2025DPFlowAdaptiveOptical} & 1.02 & 2.26 & 3.37 & \textbf{11.1} & 414 \\
WAFT~\citep{wang2025WAFTWarpingAloneField} & 1.28 & 2.56 & 3.49 & 12.9 & 732 \\
\textbf{\sname{}} & 1.03 & \textbf{2.10} & \textbf{3.11} & 11.9 & 796 \\
\bottomrule
\end{tabular}
\end{table}

\providecommand{\qGraphicsOptions}{}
\providecommand{\qFull}[9]{\begin{tikzpicture}
\node[anchor=south west,inner sep=0pt] (im) at (0,0) {\includegraphics[width=\linewidth]{#1}};
\panelframe{im}
\begin{scope}[x={(im.south east)},y={(im.north west)}]
\croplocator[#9]{cropContext}{{#4/#2},{1-#5/#3}}{{(#4+#6)/#2},{1-(#5+#7)/#3}}
\end{scope}
\if\relax\detokenize{#8}\relax\else
\node[anchor=north west,inner sep=1pt,fill=white,fill opacity=.8,text opacity=1,font=\tiny] at ([xshift=1pt,yshift=-1pt]im.north west) {#8};
\fi
\end{tikzpicture}}
\providecommand{\qCropMark}[8]{\begingroup
\pgfmathsetlengthmacro{\qImageW}{\linewidth*(#2/#6)}
\pgfmathsetlengthmacro{\qCropH}{\linewidth*(#7/#6)}
\pgfmathsetlengthmacro{\qShiftX}{-\linewidth*(#4/#6)}
\pgfmathsetlengthmacro{\qShiftY}{-\linewidth*((#3-#5-#7)/#6)}
\begin{tikzpicture}
\path[use as bounding box] (0,0) rectangle (\linewidth,\qCropH);
\begin{scope}[overlay]
\clip (0,0) rectangle (\linewidth,\qCropH);
\node[anchor=south west,inner sep=0pt,outer sep=0pt] (im) at (\qShiftX,\qShiftY) {\expandafter\includegraphics\expandafter[\qGraphicsOptions,width=\qImageW]{#1}};
\node[inner sep=0pt,outer sep=0pt,anchor=south west,minimum width=\linewidth,minimum height=\qCropH] (qpanel) at (0,0) {};
\panelframe{qpanel}
#8
\end{scope}
\end{tikzpicture}\endgroup
}
\providecommand{\qCrop}[7]{\qCropMark{#1}{#2}{#3}{#4}{#5}{#6}{#7}{}}

\providecommand{\qCropDetail}[8][]{{\def\qGraphicsOptions{interpolate=false}\qCropMark{#2}{#3}{#4}{#5}{#6}{#7}{#8}{\if\relax\detokenize{#1}\relax\else
\node[anchor=south west,inner sep=0pt,outer sep=0pt] at (im.south west) {\includegraphics[width=\qImageW]{#1}};\fi
\cropframe{qpanel}{cropContext}}}}

\begin{figure}[!t]
\centering\setlength{\parskip}{0pt}
\begingroup\offinterlineskip
\noindent\hbox{\begin{minipage}[b]{\dimexpr(\linewidth-5pt)/6\relax}\centering\scriptsize Image 1\end{minipage}\hspace{1pt}\begin{minipage}[b]{\dimexpr(\linewidth-5pt)/6\relax}\centering\scriptsize Image 2\end{minipage}\hspace{1pt}\begin{minipage}[b]{\dimexpr(\linewidth-5pt)/6\relax}\centering\scriptsize Target visibility\end{minipage}\hspace{1pt}\begin{minipage}[b]{\dimexpr(\linewidth-5pt)/6\relax}\centering\scriptsize WAFT\end{minipage}\hspace{1pt}\begin{minipage}[b]{\dimexpr(\linewidth-5pt)/6\relax}\centering\scriptsize \sname{}\end{minipage}\hspace{1pt}\begin{minipage}[b]{\dimexpr(\linewidth-5pt)/6\relax}\centering\scriptsize Ground truth\end{minipage}}\par\vspace{1pt}
\noindent\hbox{\begin{minipage}[c]{\dimexpr(\linewidth-5pt)/6\relax}
\centering
\qCrop{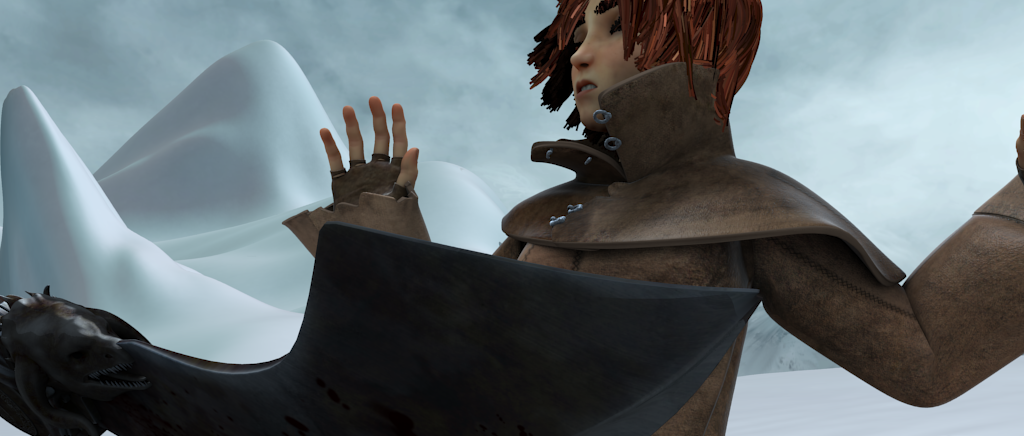}{1024}{436}{32}{0}{872}{436}
\end{minipage}\hspace{1pt}\begin{minipage}[c]{\dimexpr(\linewidth-5pt)/6\relax}
\centering
\qCrop{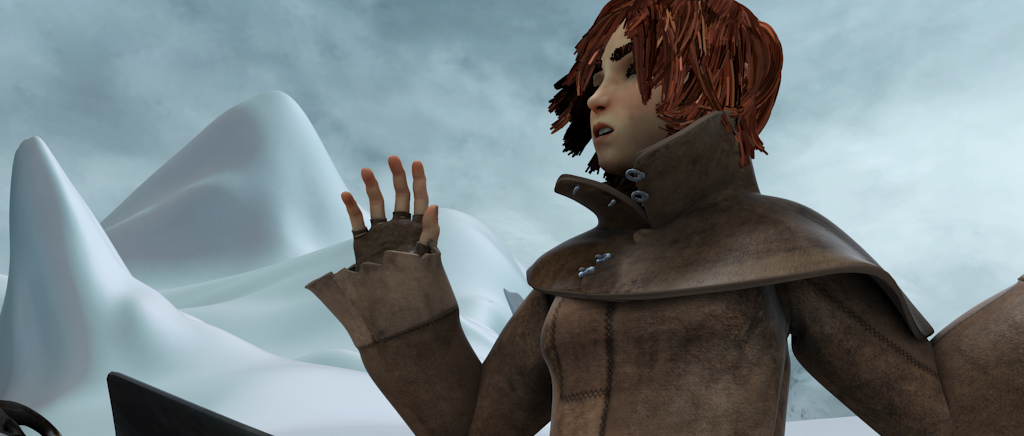}{1024}{436}{32}{0}{872}{436}
\end{minipage}\hspace{1pt}\begin{minipage}[c]{\dimexpr(\linewidth-5pt)/6\relax}
\centering
\qCrop{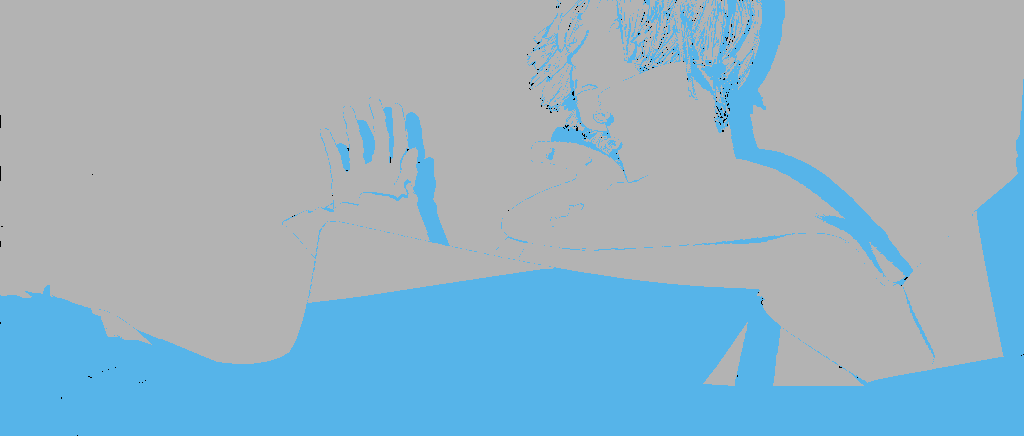}{1024}{436}{32}{0}{872}{436}
\end{minipage}\hspace{1pt}\begin{minipage}[c]{\dimexpr(\linewidth-5pt)/6\relax}
\centering
\qCrop{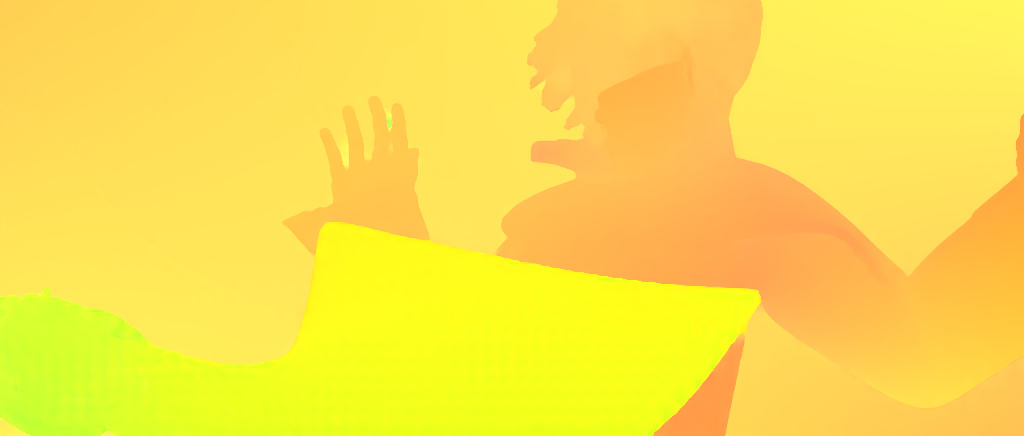}{1024}{436}{32}{0}{872}{436}
\end{minipage}\hspace{1pt}\begin{minipage}[c]{\dimexpr(\linewidth-5pt)/6\relax}
\centering
\qCrop{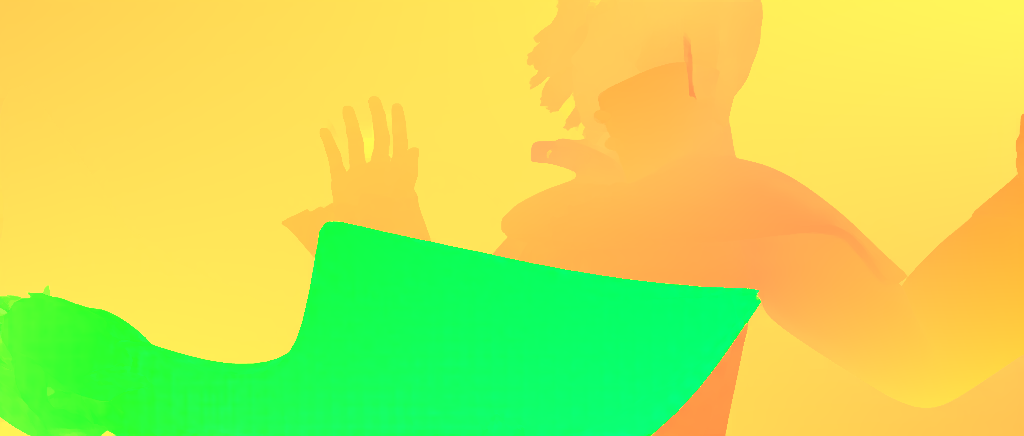}{1024}{436}{32}{0}{872}{436}
\end{minipage}\hspace{1pt}\begin{minipage}[c]{\dimexpr(\linewidth-5pt)/6\relax}
\centering
\qCrop{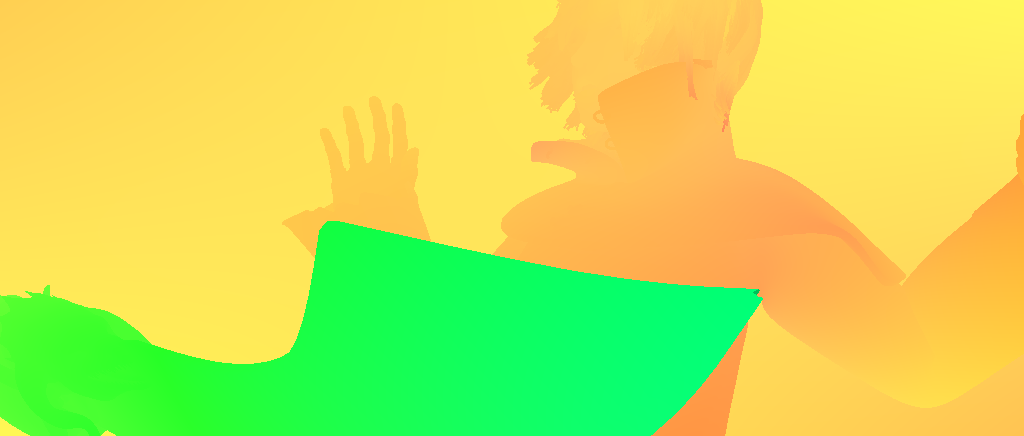}{1024}{436}{32}{0}{872}{436}
\end{minipage}}\par\vspace{2pt}
\noindent\hbox{\begin{minipage}[c]{\dimexpr(\linewidth-5pt)/6\relax}
\centering
\qCrop{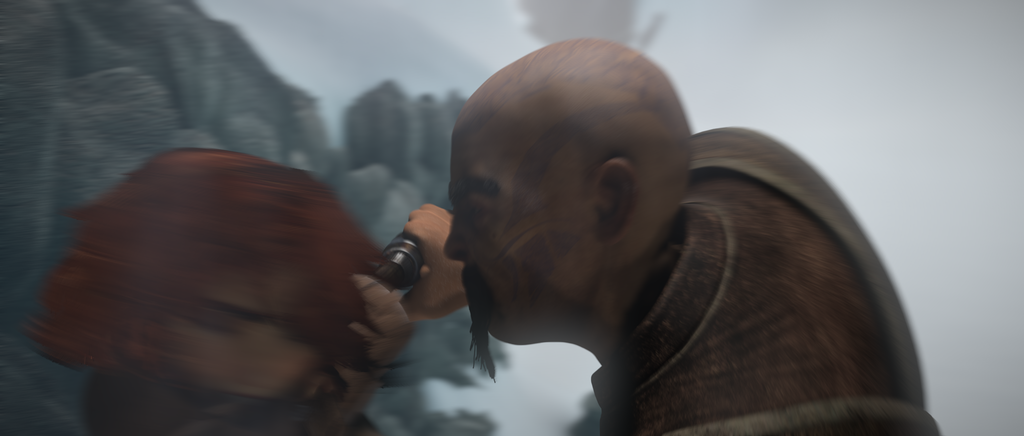}{1024}{436}{0}{0}{872}{436}
\end{minipage}\hspace{1pt}\begin{minipage}[c]{\dimexpr(\linewidth-5pt)/6\relax}
\centering
\qCrop{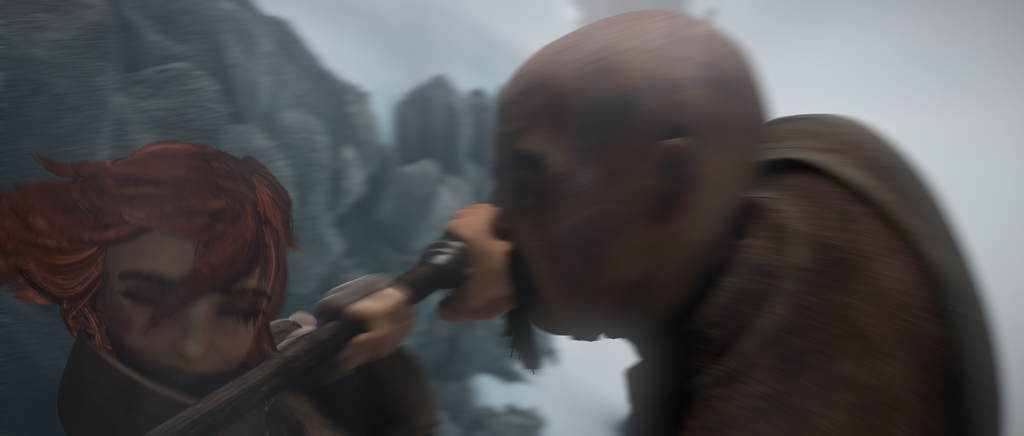}{1024}{436}{0}{0}{872}{436}
\end{minipage}\hspace{1pt}\begin{minipage}[c]{\dimexpr(\linewidth-5pt)/6\relax}
\centering
\qCrop{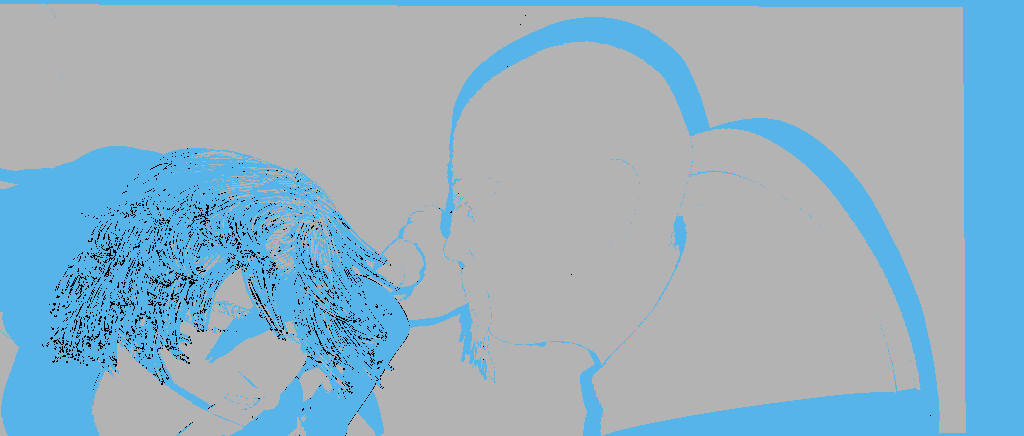}{1024}{436}{0}{0}{872}{436}
\end{minipage}\hspace{1pt}\begin{minipage}[c]{\dimexpr(\linewidth-5pt)/6\relax}
\centering
\qCrop{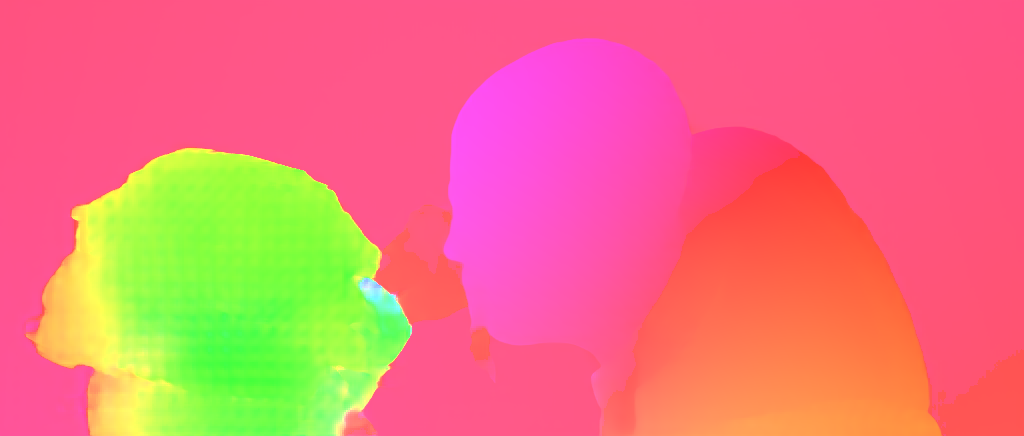}{1024}{436}{0}{0}{872}{436}
\end{minipage}\hspace{1pt}\begin{minipage}[c]{\dimexpr(\linewidth-5pt)/6\relax}
\centering
\qCrop{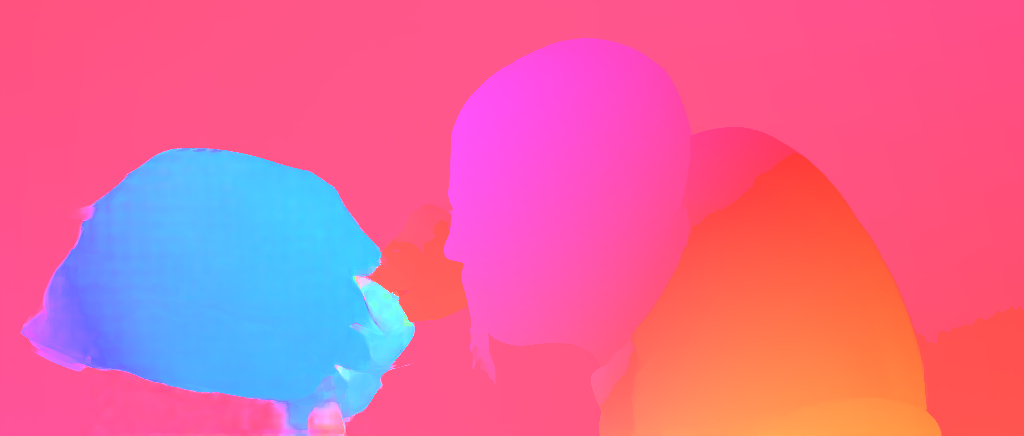}{1024}{436}{0}{0}{872}{436}
\end{minipage}\hspace{1pt}\begin{minipage}[c]{\dimexpr(\linewidth-5pt)/6\relax}
\centering
\qCrop{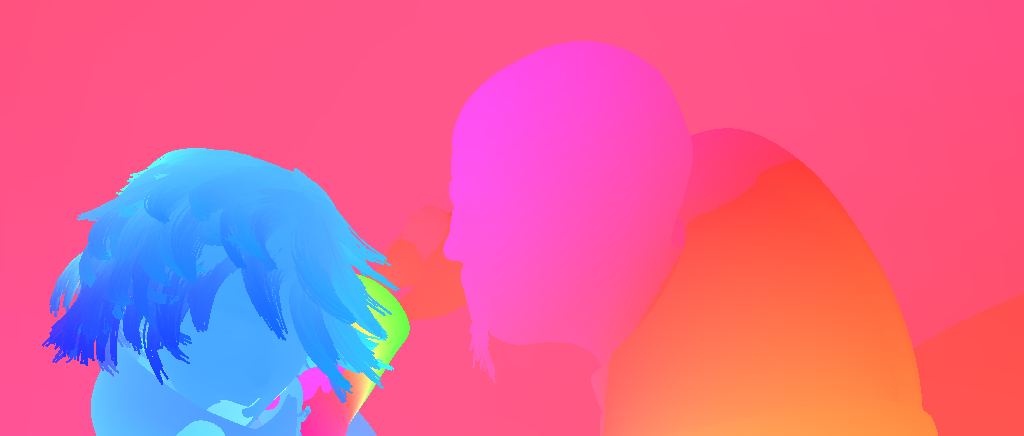}{1024}{436}{0}{0}{872}{436}
\end{minipage}}\par\vspace{2pt}
\noindent\hbox{\begin{minipage}[c]{\dimexpr(\linewidth-5pt)/6\relax}
\centering
\qCrop{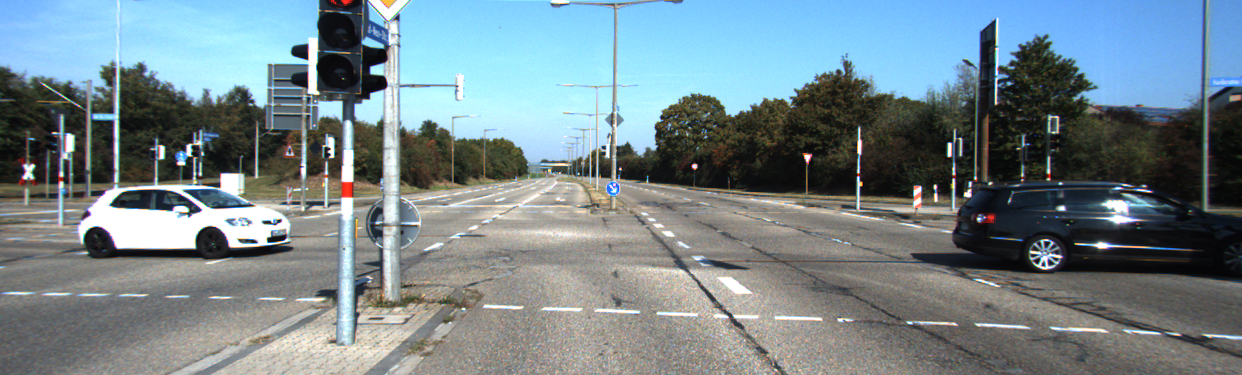}{1242}{375}{762}{125}{480}{240}
\end{minipage}\hspace{1pt}\begin{minipage}[c]{\dimexpr(\linewidth-5pt)/6\relax}
\centering
\qCrop{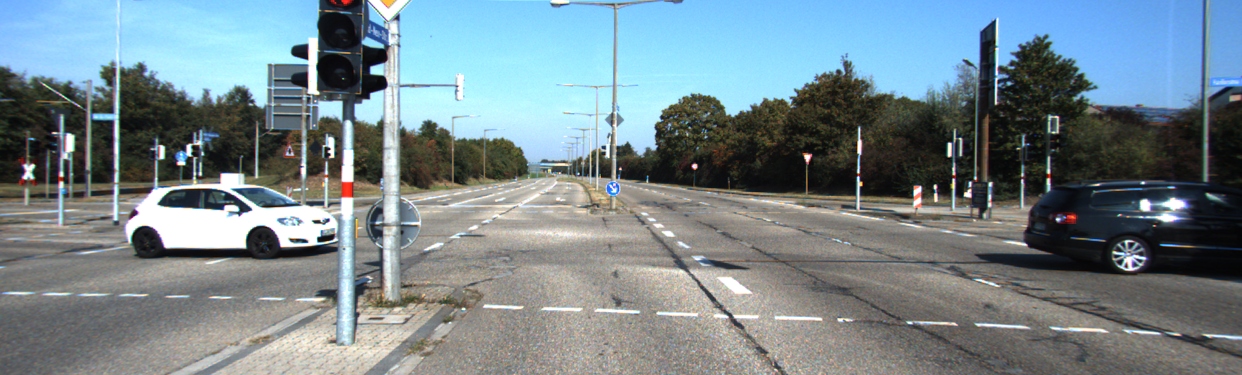}{1242}{375}{762}{125}{480}{240}
\end{minipage}\hspace{1pt}\begin{minipage}[c]{\dimexpr(\linewidth-5pt)/6\relax}
\centering
\qCrop{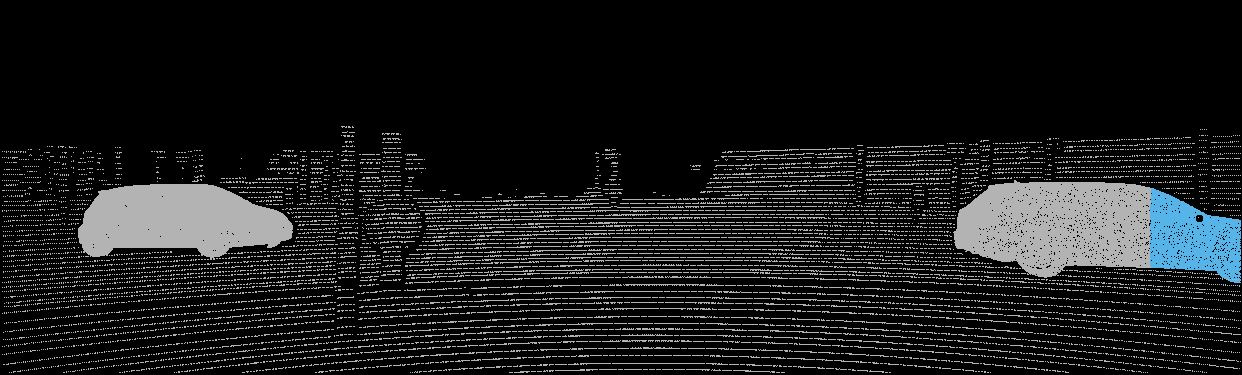}{1242}{375}{762}{125}{480}{240}
\end{minipage}\hspace{1pt}\begin{minipage}[c]{\dimexpr(\linewidth-5pt)/6\relax}
\centering
\qCrop{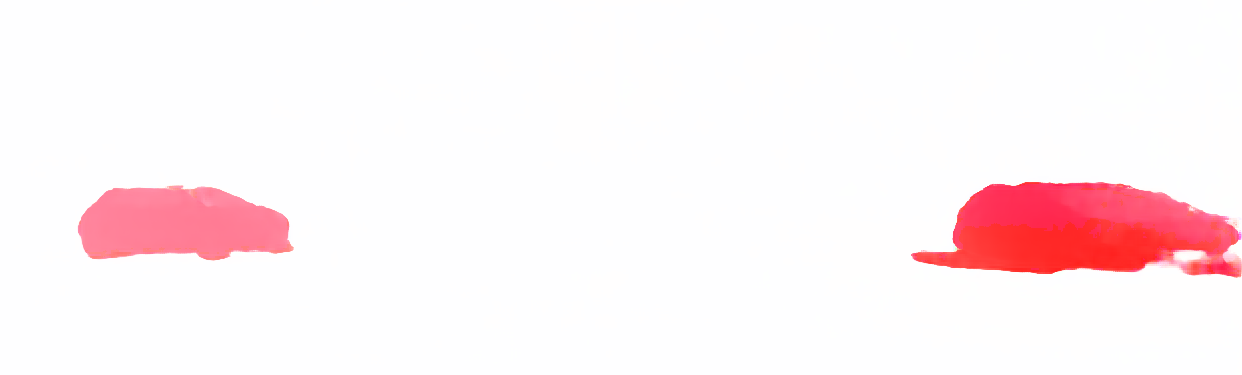}{1242}{375}{762}{125}{480}{240}
\end{minipage}\hspace{1pt}\begin{minipage}[c]{\dimexpr(\linewidth-5pt)/6\relax}
\centering
\qCrop{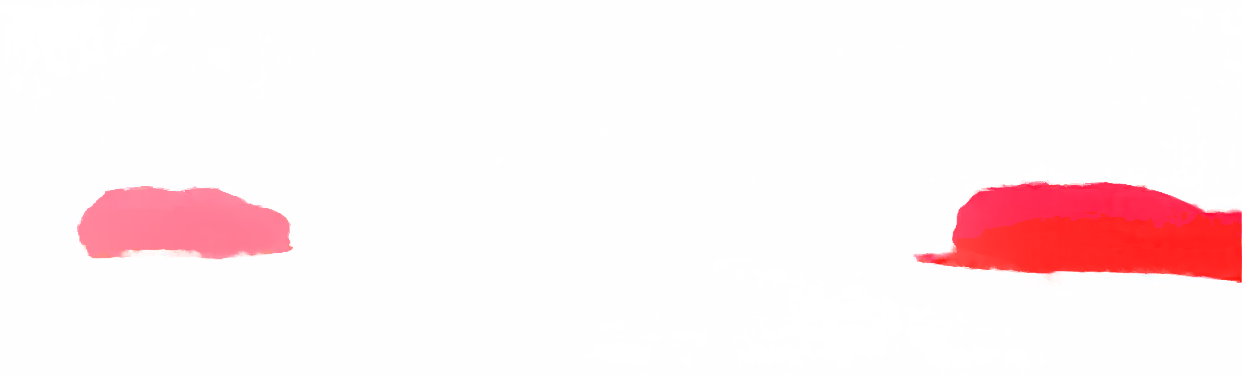}{1242}{375}{762}{125}{480}{240}
\end{minipage}\hspace{1pt}\begin{minipage}[c]{\dimexpr(\linewidth-5pt)/6\relax}
\centering
\qCrop{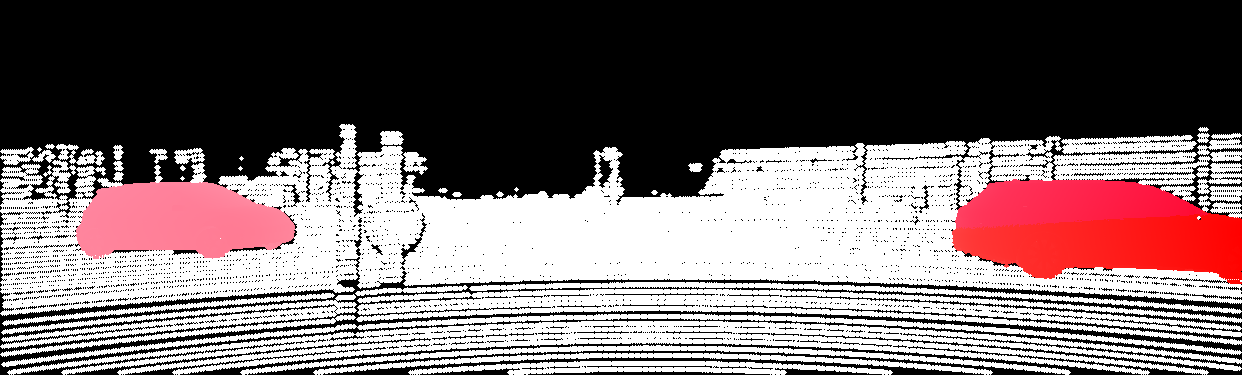}{1242}{375}{762}{125}{480}{240}
\end{minipage}}\par\vspace{2pt}
\noindent\hbox{\begin{minipage}[c]{\dimexpr(\linewidth-5pt)/6\relax}
\centering
\qCrop{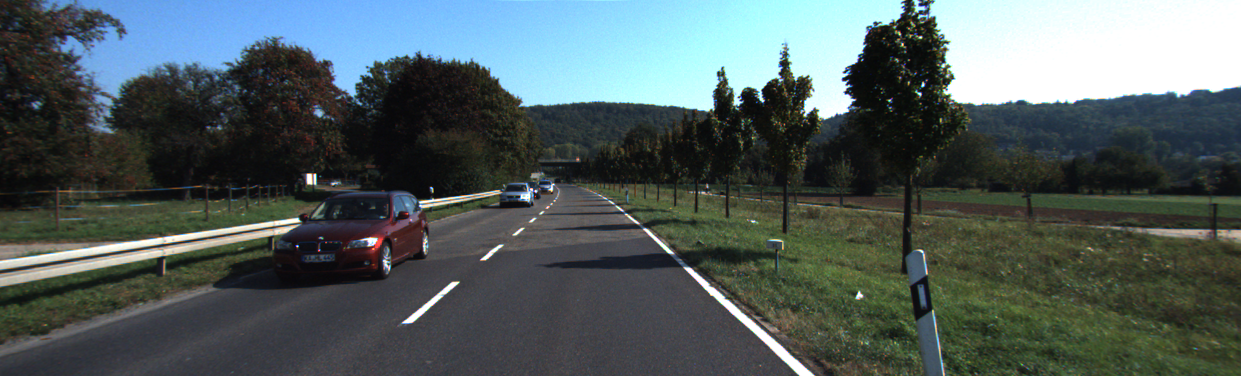}{1241}{376}{0}{76}{600}{300}
\end{minipage}\hspace{1pt}\begin{minipage}[c]{\dimexpr(\linewidth-5pt)/6\relax}
\centering
\qCrop{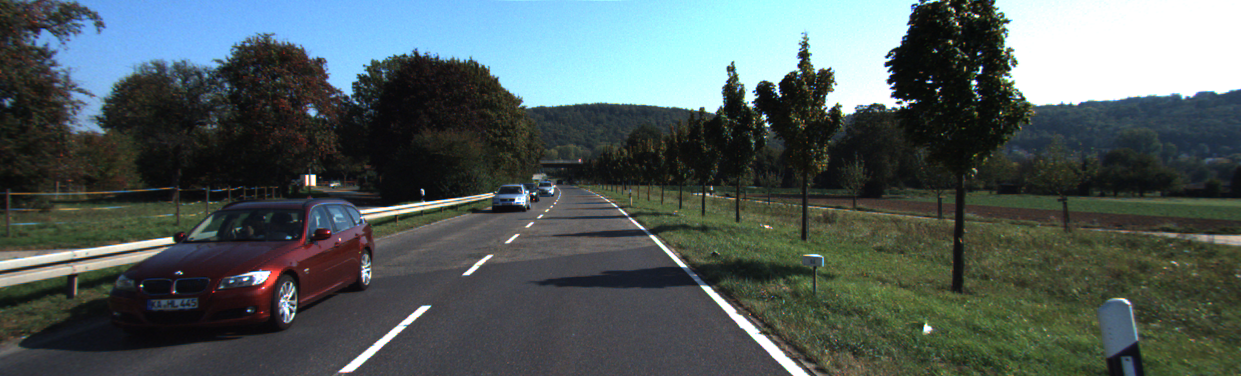}{1241}{376}{0}{76}{600}{300}
\end{minipage}\hspace{1pt}\begin{minipage}[c]{\dimexpr(\linewidth-5pt)/6\relax}
\centering
\qCrop{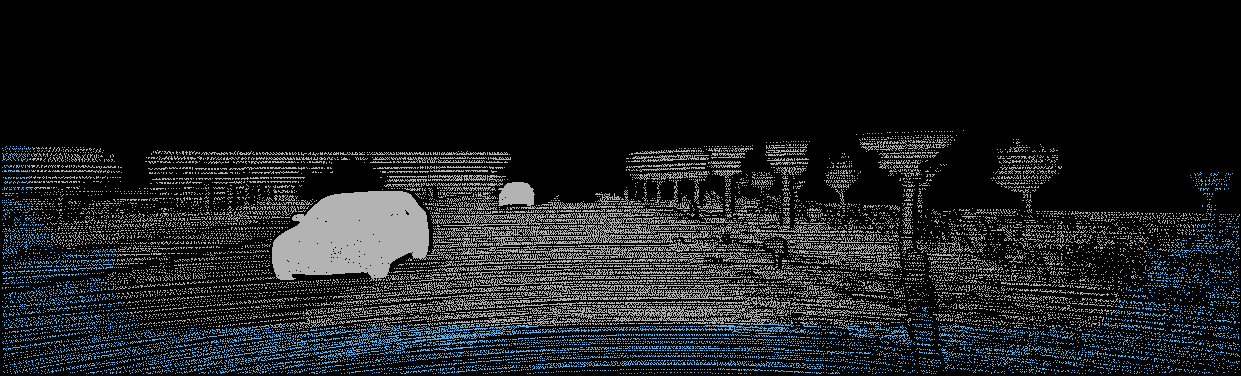}{1241}{376}{0}{76}{600}{300}
\end{minipage}\hspace{1pt}\begin{minipage}[c]{\dimexpr(\linewidth-5pt)/6\relax}
\centering
\qCrop{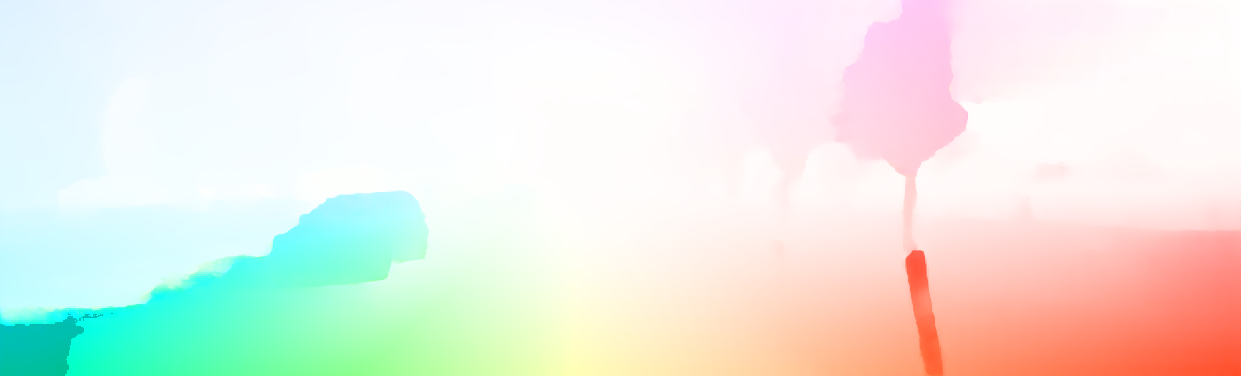}{1241}{376}{0}{76}{600}{300}
\end{minipage}\hspace{1pt}\begin{minipage}[c]{\dimexpr(\linewidth-5pt)/6\relax}
\centering
\qCrop{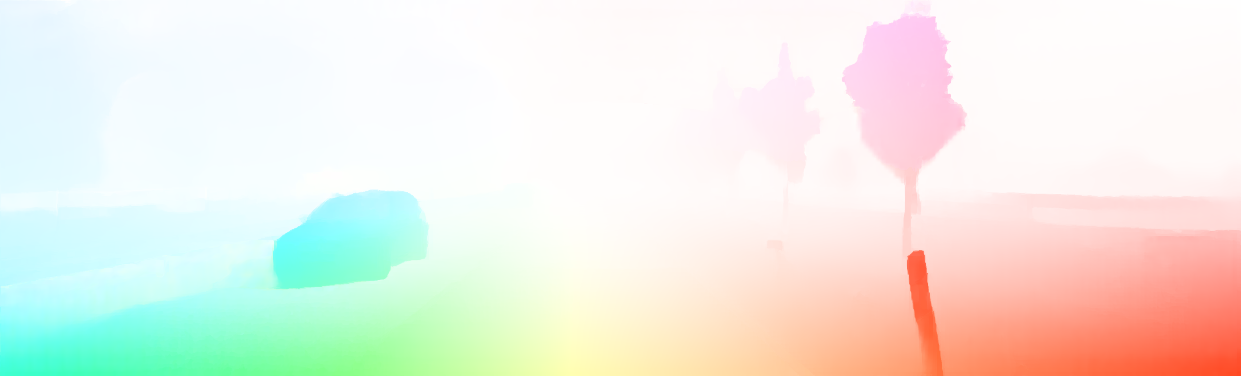}{1241}{376}{0}{76}{600}{300}
\end{minipage}\hspace{1pt}\begin{minipage}[c]{\dimexpr(\linewidth-5pt)/6\relax}
\centering
\qCrop{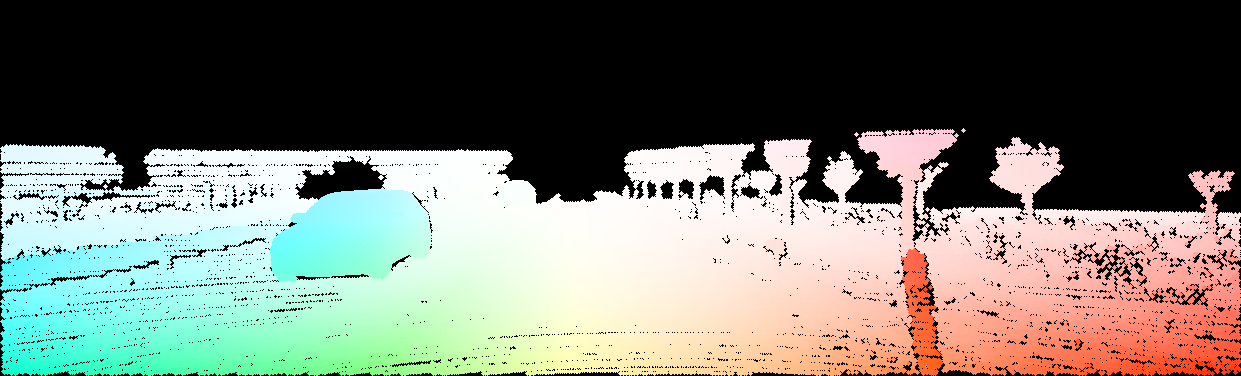}{1241}{376}{0}{76}{600}{300}
\end{minipage}}\par\endgroup
\caption{\textbf{\sname{} recovers motion where matches are missing or ambiguous and preserves motion boundaries.} Matches are missing for an occluded cape (row~1) and a car leaving the frame (row~3), and ambiguous for blurred hair (row~2); an approaching car's motion must stay separate from the road's (row~4). Target visibility marks occluded or out-of-frame pixels in blue, other valid pixels in gray, and excluded pixels in black. All predictions are zero-shot on Sintel (rows~1--2) and KITTI (rows~3--4); each row shares a crop and color scale, and sparse KITTI labels are enlarged for display.}
\label{fig:qualitative}
\end{figure}
\vspace{-6pt}

\begin{table}[!tp]
\centering\footnotesize
\caption{\textbf{\sname{}'s gain over WAFT is larger at occluded or out-of-frame
pixels and near motion boundaries than at coherent pixels.} The categories partition
valid pixels, with visible pixels split by their distance to ground-truth motion
boundaries (Appendix~\ref{app:visibility}). Pixel and Error give percentages of the
pixels and of WAFT's summed endpoint error; WAFT and \sname{} give pooled EPE, and Gain
is WAFT minus \sname{}. Bold marks the lower EPE in each pair.}
\label{tab:visibility}
\setlength{\tabcolsep}{0pt}
\begin{tabular*}{\linewidth}{@{\extracolsep{\fill}}l*{4}{r}@{\hspace{3pt}}r@{\hspace{0.5em}}*{4}{r}@{\hspace{3pt}}r@{\hspace{0.5em}}*{4}{r}@{\hspace{3pt}}r@{}}
\toprule
& \multicolumn{5}{c}{Coherent} & \multicolumn{5}{c}{Near boundary}
& \multicolumn{5}{c}{Occluded/out-of-frame} \\
\cmidrule(lr){2-6}\cmidrule(lr){7-11}\cmidrule(l){12-16}
Dataset & Pixel & Error & WAFT & \sname{} & Gain
& Pixel & Error & WAFT & \sname{} & Gain
& Pixel & Error & WAFT & \sname{} & Gain \\
\midrule
Sintel Clean & 87.9 & 36.4 & 0.52 & \textbf{0.38} & 0.14
& 3.6 & 10.8 & 3.79 & \textbf{3.24} & 0.54
& 8.4 & 52.7 & 7.96 & \textbf{6.78} & 1.18 \\
Sintel Final & 87.9 & 46.7 & 1.35 & \textbf{1.05} & 0.29
& 3.6 & 10.7 & 7.51 & \textbf{6.63} & 0.87
& 8.4 & 42.5 & 12.88 & \textbf{10.93} & 1.95 \\
KITTI & 77.8 & 36.8 & \textbf{1.72} & 1.76 & $-$0.04
& 8.1 & 10.8 & 4.83 & \textbf{4.81} & 0.01
& 13.9 & 52.3 & 13.63 & \textbf{11.05} & 2.57 \\
\midrule
Things Clean & 84.8 & 35.4 & 1.45 & \textbf{0.92} & 0.53
& 8.7 & 22.0 & 8.82 & \textbf{5.76} & 3.06
& 6.4 & 42.5 & 23.04 & \textbf{16.56} & 6.47 \\
Things Final & 84.8 & 35.1 & 1.39 & \textbf{0.93} & 0.46
& 8.7 & 22.9 & 8.87 & \textbf{6.03} & 2.83
& 6.4 & 41.9 & 21.96 & \textbf{15.88} & 6.07 \\
\bottomrule
\end{tabular*}
\end{table}

\textit{3)~\sname{}'s largest gains occur where direct matches are missing.}\enskip
Coherent pixels gain at most 0.53 pixels, versus 1.18--2.57 pixels for
occluded or out-of-frame pixels on Sintel and KITTI and over 6 on Things,
the largest gain in each row of Table~\ref{tab:visibility}.
Without direct matches, these pixels' motion must be inferred from shared
evidence, such as visible parts of the same moving region.
In Fig.~\ref{fig:regions}(b), \sname{} recovers the motion of leaves
hidden in Image~2, unlike WAFT; Fig.~\ref{fig:qualitative} shows
similar cases in rows~1 and~3. Each token pools evidence over its
learned support, draws on local context, and contributes its motion
estimate to associated pixels; Table~\ref{tab:h4}(b) tests the benefit of context at equal
token counts. Since these pixels are few, overall EPE dilutes their
gain to 0.38 pixels on KITTI (Table~\ref{tab:zero-shot}), suggesting
greater benefit where missing matches are more common.

\textit{4)~\sname{} is more precise near motion boundaries with a quarter of WAFT's tokens.}\enskip
Mixing motions from different surfaces can give boundary pixels
incorrect flow, a risk finer tokens should reduce. Yet \sname{}'s
near-boundary EPE is 0.54--3.06 pixels lower than WAFT's on Sintel
and Things, three to six times its coherent gain, and matches WAFT
on KITTI. In Fig.~\ref{fig:regions}(a), WAFT assigns a thin protrusion
the background's motion, whereas \sname{} keeps it distinct.
Including occluded or out-of-frame pixels, \sname{} lowers boundary
EPE on every dataset at all three thresholds
(Appendix~\ref{app:boundaries}; Fig.~\ref{fig:qualitative}, row~4).
As Figs.~\ref{fig:teaser-analysis} and~\ref{fig:structure} show,
learned support follows moving surfaces across grid-cell borders,
separating pixels on opposite sides of motion boundaries.

\paragraph{Adaptive versus fixed-grid correspondence.}
\sname{} differs from released WAFT in the number of tokens and how they
are formed. To remove the first difference, we train WAFT-H/4, a fixed-grid
baseline that estimates flow on an $H/4$ grid with WAFT's code and loss. It
matches \sname{}'s token count, backbone, dense-feature resolution, iterative
framework, refinement steps and training recipe, but its tokens are
fixed grid cells. We train both with and without local attention and compare
accuracy, the use of context for pixels without direct matches and the
preservation of pixel motion.

\begin{table}[!tp]
\centering\footnotesize\caption{\textbf{At the same token count, adaptive tokens improve accuracy, gain more from local context and preserve more pixel motion.}
All systems estimate flow for $H/4$ tokens.
(a) Overall accuracy. (b) Occluded/out-of-frame EPE; attention gain is EPE without minus with local attention.
(c) EPE of ground-truth flow pooled into tokens and returned to pixels with each model's image-predicted weights, without retraining.
Bold marks the best in (a) and (c) and the larger gain in (b).}
\label{tab:h4}\label{tab:attention-support}\label{tab:support}
\setlength{\tabcolsep}{2pt}\renewcommand{\arraystretch}{1.03}\begin{minipage}[t]{.49\linewidth}\vspace{0pt}\centering\fitwidth{\begin{tabular}{lcrrrrrr}
\toprule
\multicolumn{2}{l}{\textit{(a) Overall accuracy}}&
\multicolumn{2}{c}{Things} & \multicolumn{2}{c}{Sintel} & \multicolumn{2}{c}{KITTI} \\
\cmidrule(lr){3-4}\cmidrule(lr){5-6}\cmidrule(lr){7-8}
System & Attn. & Clean & Final & Clean & Final & EPE & Fl (\%) \\
\midrule
WAFT-H/4 & -- & 2.93 & 2.72 & 1.08 & 2.49 & 3.96 & 13.3 \\
WAFT-H/4 & $\checkmark$ & 2.73 & 2.62 & 1.07 & 2.57 & 3.75 & 12.6 \\
\addlinespace[2pt]
\sname{} & -- & 2.56 & 2.48 & 1.12 & 2.14 & 3.54 & 13.0 \\
\sname{} & $\checkmark$ & \textbf{2.39} & \textbf{2.38} & \textbf{1.03} & \textbf{2.10} & \textbf{3.11} & \textbf{11.9} \\
\midrule
\multicolumn{8}{l}{\textit{(c) Supplied-motion reconstruction EPE}} \\
WAFT-H/4 & $\checkmark$ & 0.50 & 0.53 & 0.21 & 0.39 & 1.52 & -- \\
\sname{} & $\checkmark$ & \textbf{0.19} & \textbf{0.21} & \textbf{0.09} & \textbf{0.19} & \textbf{0.39} & -- \\
\bottomrule
\end{tabular}}\end{minipage}\hfill\begin{minipage}[t]{.49\linewidth}\vspace{0pt}\centering\fitwidth{\begin{tabular}{lcrrrrr}
\toprule
\multicolumn{7}{l}{\textit{(b) Occluded/out-of-frame EPE}} \\
& & \multicolumn{2}{c}{Things} & \multicolumn{2}{c}{Sintel} & KITTI \\
\cmidrule(lr){3-4}\cmidrule(lr){5-6}
System & Attn. & Clean & Final & Clean & Final & EPE \\
\midrule
WAFT-H/4 & -- & 19.65 & 17.79 & 7.15 & 12.91 & 14.87 \\
WAFT-H/4 & $\checkmark$ & 18.98 & 17.64 & 7.19 & 12.90 & 14.09 \\
Attention gain &  & 0.67 & 0.15 & $-$0.03 & 0.01 & 0.78 \\
\midrule
\sname{} & -- & 17.49 & 16.55 & 6.89 & 11.40 & 12.96 \\
\sname{} & $\checkmark$ & 16.56 & 15.88 & 6.78 & 10.93 & 11.05 \\
Attention gain &  & \textbf{0.93} & \textbf{0.66} & \textbf{0.11} & \textbf{0.46} & \textbf{1.90} \\
\bottomrule
\end{tabular}}\end{minipage}

\end{table}

\textit{1)~Adaptive tokens improve every reported accuracy result.}\enskip In Table~\ref{tab:h4}(a), \sname{} outperforms both WAFT-H/4 variants
on every reported accuracy result. With attention in both systems, it
lowers Sintel Final EPE from 2.57 to 2.10 and KITTI EPE from 3.75 to 3.11.
Without attention in either, \sname{} still improves every result except Sintel
Clean EPE. \sname{}'s advantage over WAFT therefore does not come from using
coarser tokens or from local attention alone.

\textit{2)~Adaptive tokens benefit more from local context.}\enskip In Table~\ref{tab:h4}(b), \sname{} has lower occluded/out-of-frame EPE than
WAFT-H/4 even without local attention, and adding attention widens this gap on
every dataset. Attention lowers \sname{}'s EPE by 1.90 pixels on
KITTI versus 0.78 for WAFT-H/4, and changes WAFT-H/4's EPE on Sintel by less
than 0.05 pixels. This supports our account of the gains in
Table~\ref{tab:visibility}: learned support defines which pixels
share an estimate, so a token can draw on broader context while its estimate
remains tied to the pixels it represents.

\textit{3)~Learned support preserves more pixel motion.}\enskip Grouping pixels to share evidence should not erase the motion differences among
them. To measure how much pixel motion survives grouping, independent of
matching accuracy, we replace predicted token flow with pooled ground truth and
return it to pixels with each model's image-predicted weights
(Appendix~\ref{app:support}). In Table~\ref{tab:h4}(c), \sname{} has lower
reconstruction EPE than WAFT-H/4 on every dataset: 0.19 versus 0.39 on
Sintel Final and 0.39 versus 1.52 on KITTI. At the same token count, learned
support thus preserves motion that fixed cells average away. Consistent with
this, \sname{} also has lower EPE than WAFT-H/4 near motion boundaries on
every dataset at all three tested thresholds (Appendix~\ref{app:boundaries}).

\paragraph{Learning effective correspondence tokens.}
Table~\ref{tab:components} tests how supervision, local context and the
assignment parameterization help tokens learn, changing one choice at a
time and training each with the same recipe. Each design choice
improves at least five of the six reported results.

\begin{table}[!t]
\centering
\footnotesize
\caption{\textbf{Each design choice improves at least five of the six results.}
Each variant changes one choice of \sname{} and is trained with the same
recipe. \emph{No reconstruction loss} trains without Eq.~\eqref{eq:flowloss}. \emph{No local attention} forms token features by a
learned projection of pooled features. \emph{End-to-end assignment supervision}
lets the assignment loss (Eq.~\eqref{eq:mixture}) also update token flow and
uncertainty; pseudo-ground-truth token flow stays fixed. With \emph{independent assignment
scores}, the assignment head predicts the nine candidate scores directly
instead of through a Gaussian. Errors are EPE except KITTI Fl. Bold marks the
lowest error in each column.}
\label{tab:components}
\setlength{\tabcolsep}{3pt}
\begin{tabular*}{\linewidth}{@{\extracolsep{\fill}}lcccccc@{}}
\toprule
Variant & Things Clean & Things Final & Sintel Clean & Sintel Final & KITTI EPE & KITTI Fl (\%) \\
\midrule
\sname{} & \textbf{2.39} & \textbf{2.38} & 1.03 & \textbf{2.10} & \textbf{3.11} & \textbf{11.9} \\
No reconstruction loss & 2.68 & 2.61 & 1.05 & 2.17 & 3.14 & 12.4 \\
No local attention & 2.56 & 2.48 & 1.12 & 2.14 & 3.54 & 13.0 \\
End-to-end assignment supervision & 2.49 & 2.46 & \textbf{1.00} & 2.18 & 3.64 & 12.3 \\
Independent assignment scores & 2.58 & 2.55 & 1.02 & 2.24 & 3.16 & 12.1 \\
\bottomrule
\end{tabular*}
\end{table}

\textit{1)~Training the grouping to preserve motion improves every result.}\enskip Compared with training without it, the reconstruction loss
(Eq.~\eqref{eq:flowloss}) improves every result, most on Things, where Clean
EPE falls from 2.68 to 2.39. Unlike the token and assignment losses, which act
through predicted token flow, it trains pooling and broadcasting directly on
ground-truth flow, so the grouping learns which motion differences to preserve independently
of matching accuracy.

\textit{2)~Local attention improves every result.}\enskip Compared with a learned projection of pooled features, local attention improves
every result, most on KITTI, where EPE falls from 3.54 to 3.11. It lets each
token draw on nearby dense features, while the assignments still decide which
pixels receive its flow.

\textit{3)~Supervising assignments directly, rather than end to end, improves accuracy.}\enskip \sname{} supervises each module for its role (Fig.~\ref{fig:unit-spaces}).
The token loss trains matching to produce each token's correspondence, and the
assignment loss trains the assignments to decide which token should carry each
pixel, given the current probabilistic token-flow predictions. Applied end to
end, the assignment loss also updates these predictions, so it can be lowered
by shifting a token's flow or widening its uncertainty toward a disagreeing
pixel instead of reassigning that pixel. Direct assignment supervision improves
every result except Sintel Clean EPE, most on KITTI, where EPE falls from 3.64
to 3.11. Features, token predictions and assignments are still learned jointly.

\textit{4)~Gaussian assignment scores outperform independently predicted scores.}\enskip Compared with predicting the nine candidate scores independently from the same head,
Gaussian scores improve every result except Sintel Clean EPE, including Sintel
Final EPE, which falls from 2.24 to 2.10. The Gaussian couples each pixel's
candidate scores through the anchors' relative positions
(Appendix~\ref{app:method-details}), yet a token's support can still take an
irregular shape (Fig.~\ref{fig:structure}).

\paragraph{Accuracy and computation.}
\sname{} predicts assignments at image resolution but matches and refines
fewer tokens. Appendix~\ref{app:compute} profiles how this
allocation affects accuracy and cost.

\textit{1)~Fewer tokens make wider refinement and local attention affordable.}\enskip In Table~\ref{tab:compute}, with a quarter as many tokens, \sname{} refines with twice WAFT's
recurrent-state and refinement width at similar refinement cost, 453.9 versus
425.8 GMACs, whereas widening WAFT the same way would more than double its
refinement cost. Local attention of the same width
likewise costs 20.6 GMACs over \sname{}'s $H/4$ tokens versus 57.0 GMACs over WAFT's
$H/2$ grid.

\textit{2)~With four refinement steps, \sname{} outperforms five-step WAFT at lower cost.}\enskip In Table~\ref{tab:iterations}, four-step \sname{} has lower error than five-step WAFT on every reported result
while using 706.1 versus 732.4 GMACs. Our
five-step default costs more than WAFT, 796.8 GMACs, and further lowers error on
five of the six reported results. A narrower \sname{} with WAFT's width costs
636.3 GMACs and has lower error than WAFT on every reported result except KITTI
EPE.

\section{Conclusion}

Optical flow requires reliable matching from sufficient visual
context and precise motion at every pixel.
\sname{} reconciles these demands by decoupling correspondence scale
from flow resolution while maintaining consistent visual grounding.
Learned pixel-to-token associations adapt the size and shape of
correspondence support to image structure, pooling features for
matching and combining token motions into dense pixel flow.
Flow supervision learns these associations without grouping labels,
allowing the model to predict them from appearance alone.

\sname{} achieves the lowest zero-shot Sintel Final and KITTI EPE
among compared published methods. Its largest gains occur at
occluded or out-of-frame pixels, with further gains near
motion boundaries. At equal token counts, adaptive grouping
outperforms fixed-grid alternatives, benefits more from local attention,
and better preserves pixel motion.
With four refinement steps, \sname{} also outperforms five-step WAFT
at lower cost.
These results show that adaptive correspondence support enables
reliable matching and precise pixel-level flow, with strong
generalization to unseen real scenes.

\label{main-text-end}
\clearpage
\section*{Acknowledgements}
This project was supported in part by NSF grant 2313151, with additional computing resources provided through NAIRR Pilot CIS240421 and CIS240431.

\clearpage
\appendix
\section*{Appendix Outline}
The appendix is organized as follows.
\begin{itemize}
\item Appendix~\ref{app:backbone}: \nameref{app:backbone}.
\item Appendix~\ref{app:visibility}: \nameref{app:visibility}.
\item Appendix~\ref{app:support}: \nameref{app:support}.
\item Appendix~\ref{app:compute}: \nameref{app:compute}.
\item Appendix~\ref{app:finetuning}: \nameref{app:finetuning}.
\item Appendix~\ref{app:method-details}: \nameref{app:method-details}.
\end{itemize}

\noindent\textbf{Common setting.}
Unless stated otherwise, we evaluate the DINOv3 systems from
Section~\ref{sec:experiments}: \sname{}, released WAFT, and the fixed-grid
WAFT-H/4 baseline with local attention, all trained on FlyingChairs and
FlyingThings3D. The fine-tuned results in Appendix~\ref{app:finetuning} are the
exception.

\section{Zero-shot transfer across backbones}
\label{app:backbone}
In Table~\ref{tab:backbones}, \sname{} lowers zero-shot EPE on Sintel and KITTI
with both DINOv3 and Depth Anything V2 (DAv2)~\citep{yang2024DepthAnythingV2},
compared with WAFT's published results~\citep{wang2025WAFTWarpingAloneField}.
KITTI Fl also improves with DINOv3, and with DAv2 the two models tie at the
reported precision.
\begin{table}[H]
\centering
\small
\caption{\textbf{\sname{} lowers EPE with both backbones.} Zero-shot results on the
Sintel and KITTI training splits. Bold: lowest error within each backbone,
including ties at the reported precision.}
\label{tab:backbones}
\setlength{\tabcolsep}{6pt}
\renewcommand{\arraystretch}{1.08}
\begin{tabular}{llrrrr}
\toprule
& & \multicolumn{2}{c}{Sintel} & \multicolumn{2}{c}{KITTI} \\
\cmidrule(lr){3-4}\cmidrule(lr){5-6}
Backbone & Method & Clean & Final & EPE & Fl (\%) \\
\midrule
DINOv3 & WAFT~\citep{wang2025WAFTWarpingAloneField} & 1.28 & 2.56 & 3.49 & 12.9 \\
 & \sname{} & \textbf{1.03} & \textbf{2.10} & \textbf{3.11} & \textbf{11.9} \\
\midrule
DAv2 & WAFT~\citep{wang2025WAFTWarpingAloneField} & 1.01 & 2.49 & 3.28 & \textbf{10.9} \\
 & \sname{} & \textbf{0.97} & \textbf{2.30} & \textbf{2.93} & \textbf{10.9} \\
\bottomrule
\end{tabular}
\end{table}

\section{Regional evaluation}
\label{app:visibility}
The regional analyses test accuracy where direct matches are unavailable, at
occluded or out-of-frame pixels, and where neighboring motions differ, near
motion boundaries.

\subsection{Occluded and out-of-frame pixels}
\label{app:protocol}
\label{app:error-burden}
\paragraph{Pixel sets.}
A pixel is out of frame if its ground-truth endpoint $x+F_x^{\mathrm{gt}}(x)$ falls outside
Image~2. On Sintel and KITTI, we combine these pixels with those whose targets
are marked occluded by the dataset annotations. On Things, we use endpoint
bounds only, so in-frame occlusions are not identified and visible means in
frame. All sets are restricted to the valid pixels used for standard EPE.

Sintel's official invalid mask further excludes pixels from visibility
analysis, even when they remain valid for standard EPE. Table~\ref{tab:visibility}
partitions the remaining population: visible-boundary pixels lie in the
boundary neighborhoods defined below, visible-coherent pixels are their
complement among visible pixels (coherent does not imply constant flow), and
the third group is non-visible.

\paragraph{Regional EPE and error share.}
Let $V$ be the pixels eligible for visibility analysis, pooled over all
evaluated image pairs, $R\subseteq V$ one of the three categories, and
$e(x)=\|F_x(x)-F_x^{\mathrm{gt}}(x)\|_2$ the endpoint error at pixel $x$.
Table~\ref{tab:visibility} reports the regional EPE and, as a percentage,
WAFT's error share
\begin{equation}
 \mathrm{EPE}_R=\frac{1}{|R|}\sum_{x\in R}e(x),
 \qquad
 S_R=\frac{\sum_{x\in R}e(x)}{\sum_{x\in V}e(x)}.
 \label{eq:error-share}
\end{equation}
Unlike standard EPE, which averages within each image first, both weight every
pixel equally. The pixel and error shares of the three categories therefore
each sum to 100\% before truncation.

\subsection{Accuracy near motion boundaries}
\label{app:boundaries}
\paragraph{Defining the evaluation region.}
To test accuracy where neighboring motions differ, we measure EPE over a
region we define around ground-truth motion boundaries. We identify these
boundaries from local changes in ground-truth flow $F_x^{\mathrm{gt}}$, with strength
\begin{equation}
 s(x)=\|\Delta_1 F_x^{\mathrm{gt}}(x)\|_2+\|\Delta_2 F_x^{\mathrm{gt}}(x)\|_2,
 \label{eq:boundary-strength}
\end{equation}
where each $\Delta$ takes the difference to the next pixel along one image
axis. A difference contributes only when both pixels have valid labels.
For threshold $\tau$, pixels with $s(x)>\tau$ seed the boundary region,
which contains valid pixels in a $5\times5$ window around any seed.
We pool endpoint errors over this region. We evaluate $\tau=2,3,4$ pixels;
Table~\ref{tab:visibility} uses $\tau=2$ intersected with the visible set
on Sintel and KITTI, or the in-frame set on Things.
The threshold study in Table~\ref{tab:motion-boundaries} includes both visible
and non-visible pixels. At $\tau=2$, this full boundary region covers 9.3\%
of valid pixels on Things, 6.2\% on Sintel and 13.6\% on KITTI.

\paragraph{Boundary precision improves across the tested thresholds.}
In Table~\ref{tab:motion-boundaries}, \sname{} has lower EPE in this region
than both fixed-grid models on every dataset at all three thresholds.
\begin{table}[H]
\centering
\small
\caption{\textbf{\sname{} has lower EPE near motion boundaries at all tested thresholds.}
EPE is pooled over the boundary region, including non-visible pixels; the
threshold sets how large a local flow change must be to count as a boundary.
WAFT estimates motion for four times as many tokens as \sname{}, and
WAFT-H/4 with attention for the same number.}
\label{tab:motion-boundaries}
\setlength{\tabcolsep}{5pt}
\renewcommand{\arraystretch}{1.08}
\begin{tabular}{lrrrrr}
\toprule
& \multicolumn{2}{c}{Things} & \multicolumn{2}{c}{Sintel} & KITTI \\
\cmidrule(lr){2-3}\cmidrule(lr){4-5}
System & Clean & Final & Clean & Final & EPE \\
\midrule
\multicolumn{6}{l}{\textit{(a) Boundary threshold: 2 pixels}} \\
WAFT & 13.40 & 13.14 & 7.30 & 11.92 & 9.97 \\
WAFT-H/4 + attention & 10.77 & 10.65 & 6.97 & 11.40 & 11.75 \\
\sname{} & \textbf{9.24} & \textbf{9.48} & \textbf{6.63} & \textbf{10.45} & \textbf{8.51} \\
\midrule
\multicolumn{6}{l}{\textit{(b) Boundary threshold: 3 pixels}} \\
WAFT & 12.50 & 12.37 & 7.38 & 12.52 & 12.84 \\
WAFT-H/4 + attention & 10.08 & 10.07 & 7.02 & 11.90 & 15.30 \\
\sname{} & \textbf{8.56} & \textbf{8.77} & \textbf{6.64} & \textbf{11.03} & \textbf{10.65} \\
\midrule
\multicolumn{6}{l}{\textit{(c) Boundary threshold: 4 pixels}} \\
WAFT & 12.29 & 12.26 & 7.75 & 13.38 & 15.75 \\
WAFT-H/4 + attention & 10.02 & 9.99 & 7.34 & 12.73 & 18.87 \\
\sname{} & \textbf{8.40} & \textbf{8.63} & \textbf{6.85} & \textbf{11.78} & \textbf{12.76} \\
\bottomrule
\end{tabular}
\end{table}

\section{Supplied-motion reconstruction protocol}
\label{app:support}
This diagnostic measures how much pixel motion each model preserves when flow
is pooled into tokens and returned to pixels, independent of its token-flow
accuracy. Table~\ref{tab:h4}(c) reports the results.

\paragraph{Substituting pooled ground-truth flow.}
We first predict each model's weights from the input images and keep them
fixed; neither model is retrained. For \sname{}, $P_q$ pools ground-truth flow into tokens and $P_x$
broadcasts it back through the same assignments $A$. This is the ground-truth
round trip used by reconstruction supervision (Eq.~\eqref{eq:flowloss}) and
illustrated in Fig.~\ref{fig:unit-spaces}. For WAFT-H/4 with
local attention, we average ground-truth flow within each fixed cell and
upsample it with the predicted convex weights.
Pooling and evaluation use valid ground-truth pixels. We report reconstruction
EPE, averaged within each image and then over images, rather than the robust
training penalty $\mathcal D$ in Eq.~\eqref{eq:flowloss}.

\paragraph{What the reconstruction measures.}
Lower reconstruction EPE means that less pixel motion is lost in the round
trip. The comparison evaluates pooling and the return to pixels together; it
does not isolate their effects or decompose the end-to-end prediction error.

\section{Accuracy and computation}
\label{app:compute}
Computation is measured per $540\times960$ image pair with model-specific
padding. Counts include feature extraction, token matching features, all
refinement steps and the final per-pixel flow output.

\paragraph{Fewer tokens make wider refinement and local attention affordable.}
In Table~\ref{tab:compute}, suffixes give recurrent-state and refinement
widths: WAFT-64 is released WAFT, and \sname{}-128 is our main model. At
WAFT's width, refinement over \sname{}'s tokens costs 293.8 versus 425.8
GMACs. Widening WAFT to \sname{}-128's width, profiled without training,
raises its refinement cost to 943.9 GMACs.

\begin{table}[H]
\centering
\footnotesize
\caption{\textbf{Fewer tokens make wider refinement and local attention affordable.} Resolution denotes the token lattice; suffixes give recurrent-state and refinement widths; Steps counts refinement steps. Costs are GMACs per $540\times960$ image pair; errors are EPE except KITTI Fl. Bold marks the lowest refinement cost, total cost and error in each column. $\dagger$: profiled only, not trained.}
\label{tab:compute}
\setlength{\tabcolsep}{2pt}
\renewcommand{\arraystretch}{1.08}
\begin{tabular*}{\linewidth}{@{\extracolsep{\fill}}lccrrrrrrrrr@{}}
\toprule
& & & \multicolumn{3}{c}{Cost (GMACs)} & \multicolumn{2}{c}{Things} & \multicolumn{2}{c}{Sintel} & \multicolumn{2}{c}{KITTI} \\
\cmidrule(lr){4-6}\cmidrule(lr){7-8}\cmidrule(lr){9-10}\cmidrule(lr){11-12}
System & Resolution & Steps & Attention & Refinement & Total & Clean & Final & Clean & Final & EPE & Fl (\%) \\
\midrule
WAFT-64~\citep{wang2025WAFTWarpingAloneField} & $H/2$ & 5 & -- & 425.8 & 732.4 & 3.54 & 3.44 & 1.28 & 2.56 & 3.49 & 12.9 \\
WAFT-128 + attn.$^{\dagger}$ & $H/2$ & 5 & 57.0 & 943.9 & 1313.2 & -- & -- & -- & -- & -- & -- \\
\sname{}-64 & $H/4$ & 5 & 20.6 & \textbf{293.8} & \textbf{636.3} & 2.55 & 2.48 & \textbf{1.03} & 2.24 & 3.52 & 12.5 \\
\sname{}-128 & $H/4$ & 4 & 20.6 & 363.1 & 706.1 & 2.44 & 2.43 & \textbf{1.03} & 2.15 & 3.18 & 12.2 \\
\sname{}-128 (main) & $H/4$ & 5 & 20.6 & 453.9 & 796.8 & \textbf{2.39} & \textbf{2.38} & \textbf{1.03} & \textbf{2.10} & \textbf{3.11} & \textbf{11.9} \\
\bottomrule
\end{tabular*}
\end{table}

\paragraph{Fewer refinement steps retain a favorable accuracy--computation trade-off.}
Table~\ref{tab:iterations} evaluates \sname{} and WAFT, both trained with five
refinement steps, with one to five steps at inference and no retraining. At
the same number of steps, \sname{} has lower error than WAFT on every reported
result but uses more computation. Three-step \sname{} improves on five-step
WAFT on every reported result except KITTI Fl, at 16\% lower computation
(615.3 versus 732.4 GMACs).
\begin{table}[H]
\centering
\small
\caption{\textbf{\sname{} has lower error than WAFT at every number of refinement steps.}
Both are trained with five steps and evaluated with one to five without
retraining. \sname{}-64 is separately trained with
64-dimensional recurrent-state and refinement features. Errors are EPE except
KITTI Fl.}
\label{tab:iterations}
\setlength{\tabcolsep}{5pt}
\renewcommand{\arraystretch}{1.08}
\begin{tabular}{lcrrrrrrr}
\toprule
& & \multicolumn{2}{c}{Things} & \multicolumn{2}{c}{Sintel} & \multicolumn{2}{c}{KITTI} & \\
\cmidrule(lr){3-4}\cmidrule(lr){5-6}\cmidrule(lr){7-8}
System & Steps & Clean & Final & Clean & Final & EPE & Fl (\%) & GMACs \\
\midrule
WAFT & 1 & 6.70 & 6.44 & 2.20 & 3.50 & 8.47 & 37.1 & 391.8 \\
\sname{} & 1 & 5.01 & 4.84 & 1.54 & 2.67 & 8.30 & 36.9 & 433.8 \\
\midrule
WAFT & 2 & 4.27 & 4.15 & 1.49 & 2.88 & 4.67 & 18.5 & 477.0 \\
\sname{} & 2 & 2.91 & 2.84 & 1.10 & 2.29 & 4.09 & 17.1 & 524.5 \\
\midrule
WAFT & 3 & 3.78 & 3.66 & 1.34 & 2.70 & 3.81 & 14.0 & 562.1 \\
\sname{} & 3 & 2.55 & 2.49 & 1.06 & 2.22 & 3.38 & 13.2 & 615.3 \\
\midrule
WAFT & 4 & 3.61 & 3.51 & 1.29 & 2.56 & 3.56 & 13.1 & 647.3 \\
\sname{} & 4 & 2.44 & 2.43 & 1.03 & 2.15 & 3.18 & 12.2 & 706.1 \\
\midrule
WAFT & 5 & 3.54 & 3.44 & 1.28 & 2.56 & 3.49 & 12.9 & 732.4 \\
\sname{} & 5 & 2.39 & 2.38 & 1.03 & 2.10 & 3.11 & 11.9 & 796.8 \\
\midrule
\sname{}-64 & 5 & 2.55 & 2.48 & 1.03 & 2.24 & 3.52 & 12.5 & 636.3 \\
\bottomrule
\end{tabular}
\end{table}

\section{Target-specific fine-tuning}
\label{app:finetuning}
The zero-shot advantage does not carry over to the current fine-tuned models.
Table~\ref{tab:finetuned} compares \sname{}, fine-tuned with WAFT's recipe and
submitted to the benchmark servers, with WAFT's published DINOv3
results~\citep{wang2025WAFTWarpingAloneField} on the hidden test sets of
Sintel, KITTI 2015, and Spring. \sname{} matches WAFT's reported Sintel Final
EPE and trails it on Sintel Clean, both KITTI Fl variants, and all four Spring
metrics. Extending the transfer gains to target-specific fine-tuning remains
open.

\paragraph{Benchmark metrics.}
Sintel reports EPE. KITTI reports Fl over all labeled pixels (Fl-all) and
non-occluded pixels (Fl-noc). Spring provides four ground-truth values per image pixel and
measures error against the closest one. Its 1px metric is the percentage of
pixels with endpoint error above one pixel, and WAUC summarizes accuracy
across error thresholds.
\begin{table}[H]
\centering
\small
\caption{\textbf{After fine-tuning, \sname{} matches WAFT on Sintel Final but trails it elsewhere.}
Benchmark test results for \sname{} and DINOv3 WAFT.
Lower is better except WAUC ($\uparrow$).}
\label{tab:finetuned}
\setlength{\tabcolsep}{3pt}
\renewcommand{\arraystretch}{1.08}
\begin{tabular}{lrrrrrrrr}
\toprule
& \multicolumn{2}{c}{Sintel} & \multicolumn{2}{c}{KITTI 2015} & \multicolumn{4}{c}{Spring} \\
\cmidrule(lr){2-3}\cmidrule(lr){4-5}\cmidrule(lr){6-9}
System & Clean & Final & Fl-all (\%) & Fl-noc (\%) & 1px (\%) & EPE & Fl (\%) & WAUC $\uparrow$ \\
\midrule
WAFT & 0.94 & 2.02 & 3.56 & 2.13 & 3.18 & 0.33 & 1.25 & 95.05 \\
\sname{} & 1.06 & 2.02 & 3.67 & 2.35 & 3.43 & 0.35 & 1.30 & 94.86 \\
\bottomrule
\end{tabular}
\end{table}

\section{Assignment parameterization and training objectives}
\label{app:method-details}
We specify the assignment scores and training objectives summarized in
Section~\ref{sec:method}. The operators and flow fields retain the same
pixel/token subscripts. Matrix entries select an individual pixel or token and
flow component; for example, $[F_q]_{q,c}$ is component $c$ of token $q$'s flow.
Both transformations in the loss signatures are determined by the shared assignments $A$.

\paragraph{Gaussian assignment scores.}
For pixel $x$, the anchor of its lattice cell defines the local origin.
Its candidate anchors $q\in\mathcal N(x)$ have offsets
$\delta_{xq}\in\{-1,0,1\}^2$.
The five predicted parameters are a mean offset $\mu_x\in\mathbb R^2$ and
three entries of a lower-triangular factor $L_x$ with positive diagonal,
which defines the inverse covariance $G_x=L_xL_x^\top$. The scores and
normalized assignments are
\begin{equation}
 a_{xq}=-\tfrac12(\delta_{xq}-\mu_x)^\top G_x(\delta_{xq}-\mu_x),\qquad
 A_{xq}=\frac{\exp a_{xq}}{\sum_{r\in\mathcal N(x)}\exp a_{xr}}.
 \label{eq:assignment}
\end{equation}
Entries outside $\mathcal N(x)$ are zero. Since $a_{xq}$ is the log-density
of a Gaussian with mean $\mu_x$ and covariance $G_x^{-1}$ up to a constant
that cancels in the softmax, $A_{xq}$ is that density at $\delta_{xq}$,
normalized over $\mathcal N(x)$. The parameterization couples each
pixel's candidate scores through their relative spatial positions; it does
not constrain a token's collected support to a Gaussian shape.

\paragraph{Probabilistic token-flow predictions.}
The probabilistic flow field $\mathcal F_q^{(t)}$ comprises a distribution for
each flow component of each token. We write its density as $p_{q,c}^{(t)}(f)$.
Following SEA-RAFT~\citep{wang2025SEARAFTSimpleEfficient}, it is a two-component
Laplace mixture centered on the corresponding flow estimate:
\begin{equation}
 p_{q,c}^{(t)}(f)=\sum_{k=1}^{2}\pi_{q,k}^{(t)}
 \frac{1}{2b_{q,k}^{(t)}}
 \exp\!\left(-\frac{|f-[F_q^{(t)}]_{q,c}|}{b_{q,k}^{(t)}}\right).
 \label{eq:unit-likelihood-detail}
\end{equation}
Here $c\in\{1,2\}$ indexes the flow components, and $\pi_{q,k}^{(t)}$ are
softmax-normalized mixture weights. Both mixture components share the flow
center. The implementation bounds the first log-scale to $[0,10]$ and fixes
the second to zero; scales and mixture weights are shared across the two flow
components. These mixture weights are distinct from pixel-to-token assignments.
Image inputs and model parameters are suppressed in the notation.

\paragraph{Valid labels and pseudo-ground truth.}
Let $V$ denote the valid training pixels. Write $P_q^V(A)$ for weighted pooling restricted to valid labels.
The pseudo-ground-truth token flow is $F_q^{\mathrm{gt}}=P_q^V(A)F_x^{\mathrm{gt}}$, and the validity weight
$w_q=[P_q(A)\mathbf 1_V]_q$ is the fraction of a token's support with valid labels.
The pseudo-ground truth and validity weights are recomputed from the current assignments
and held fixed for token supervision. Feature pooling and matching remain
differentiable through the assignments.

\paragraph{Mixtures of neighboring flow predictions.}
The pixelwise mixture $\operatorname{Mix}_{P_x}[\mathcal F_q^{(t)}]$ in
Eq.~\eqref{eq:mixture} has component density $p_{x,c}^{(t)}$.
Since $[P_x]_{xq}=A_{xq}$, its weights come from the pixel assignments.
For the assignment objective, let $\mathcal S(x)$ denote the candidate
tokens included in the training mixture. Their assignment weights are
renormalized within this set:
\begin{equation}
 p_{x,c}^{(t)}(f)=\sum_{q\in\mathcal S(x)}
 \frac{A_{xq}}{\sum_{r\in\mathcal S(x)}A_{xr}}\,p_{q,c}^{(t)}(f).
 \label{eq:pixel-likelihood-detail}
\end{equation}
Token-flow and uncertainty predictions are held fixed for this objective;
the assignments and their shared features learn through the mixture weights.
The end-to-end assignment supervision variant in
Table~\ref{tab:components} removes this restriction while keeping the
pseudo-ground-truth token flow fixed.
Mixing is performed separately for the two flow components, rather than
forming a likelihood around an averaged flow vector. Dense inference still
uses all nine assignment weights in $F_x=P_xF_q$.

\paragraph{Loss reductions and joint training.}
The $\operatorname{NLL}$ operator in Eqs.~\eqref{eq:unitloss} and~\eqref{eq:mixture} takes a predicted
distribution first and a target second. It averages over flow components and
aggregates over tokens or valid pixels. With pseudo-ground truth
and validity weights fixed for token supervision, and token predictions fixed
for assignment supervision, the two objectives are
\begin{align}
 \mathcal L_{\mathrm{token}}^{(t)}
 &=-\frac{1}{2\sum_q w_q}\sum_q w_q\sum_{c=1}^{2}
 \log p_{q,c}^{(t)}\!\left([F_q^{\mathrm{gt}}]_{q,c}\right),\label{eq:unitloss-detail}\\
 \mathcal L_{\mathrm{assign}}^{(t)}
 &=-\frac{1}{2|V|}\sum_{x\in V}\sum_{c=1}^{2}
 \log p_{x,c}^{(t)}\!\left([F_x^{\mathrm{gt}}]_{x,c}\right).
 \label{eq:assignloss-detail}
\end{align}
For motion preservation, we recompute
$\widetilde F_x^{\mathrm{gt}}=P_x(A)P_q^V(A)F_x^{\mathrm{gt}}$ with both transformations
differentiable. With $\rho(z)=\sqrt{\|z\|_2^2+10^{-6}}-10^{-3}$, the
reconstruction objective, specifying $\mathcal D$ in
Eq.~\eqref{eq:flowloss}, is
\begin{equation}
 \mathcal L_{\mathrm{recon}}=\frac{1}{|V|}\sum_{x\in V}
 \rho\!\left([\widetilde F_x^{\mathrm{gt}}]_{x,:}-[F_x^{\mathrm{gt}}]_{x,:}\right).
 \label{eq:flowloss-detail}
\end{equation}
This branch does not reuse the fixed pseudo-ground-truth token flow. Training
numerators and denominators pool entries across the minibatch; these training
reductions are distinct from benchmark evaluation conventions.

The first two objectives supervise every refinement step, while reconstruction
is added once:
\begin{equation}
 \mathcal L=\sum_{t=1}^{T}\gamma^{T-t}
 \left(\mathcal L_{\mathrm{token}}^{(t)}+\mathcal L_{\mathrm{assign}}^{(t)}\right)
 +\lambda_{\mathrm{recon}}\mathcal L_{\mathrm{recon}}.
 \label{eq:total}
\end{equation}
We use $T=5$, $\gamma=0.85$, and $\lambda_{\mathrm{recon}}=1$.
All objectives are optimized jointly. Holding the stated predictions fixed
within one objective separates its direct gradient paths; it does not freeze
the shared features or introduce alternating optimization.


\begin{thebibliography}{39}
\providecommand{\natexlab}[1]{#1}
\providecommand{\url}[1]{\texttt{#1}}
\expandafter\ifx\csname urlstyle\endcsname\relax
  \providecommand{\doi}[1]{doi: #1}\else
  \providecommand{\doi}{doi: \begingroup \urlstyle{rm}\Url}\fi

\bibitem[Horn and Schunck(1981)]{horn1981DeterminingOpticalFlow}
Berthold K.~P. Horn and Brian~G. Schunck.
\newblock Determining optical flow.
\newblock \emph{Artificial Intelligence}, 17\penalty0 (1):\penalty0 185--203, August 1981.
\newblock ISSN 0004-3702.
\newblock \doi{10.1016/0004-3702(81)90024-2}.

\bibitem[Lucas and Kanade(1981)]{lucas1981IterativeImageRegistration}
Bruce~D Lucas and Takeo Kanade.
\newblock An {{Iterative Image Registration Technique}} with an {{Application}} to {{Stereo Vision}}.
\newblock In \emph{{{IJCAI}}'81: 7th International Joint Conference on {{Artificial}} Intelligence}, volume~2, pages 674--679, Vancouver, Canada, August 1981.

\bibitem[Black and Anandan(1996)]{black1996RobustEstimationMultiple}
Michael~J. Black and P.~Anandan.
\newblock The robust estimation of multiple motions: Parametric and piecewise-smooth flow fields.
\newblock \emph{Computer Vision and Image Understanding}, 63\penalty0 (1):\penalty0 75--104, 1996.
\newblock URL \url{https://www.csd.uoc.gr/~hy576/bibliography/RobustMotion.pdf}.

\bibitem[Ren(2008)]{ren2008LocalGroupingOpticalFlow}
Xiaofeng Ren.
\newblock Local grouping for optical flow.
\newblock In \emph{IEEE Conference on Computer Vision and Pattern Recognition}, pages 1--8, 2008.
\newblock URL \url{https://iacl.ece.jhu.edu/proceedings/cvpr2008/papers/196.pdf}.

\bibitem[Teed and Deng(2020)]{teed2020RAFTRecurrentAllPairsa}
Zachary Teed and Jia Deng.
\newblock {{RAFT}}: {{Recurrent All-Pairs Field Transforms}} for {{Optical Flow}}.
\newblock In Andrea Vedaldi, Horst Bischof, Thomas Brox, and Jan-Michael Frahm, editors, \emph{Computer {{Vision}} -- {{ECCV}} 2020}, pages 402--419, Cham, 2020. Springer International Publishing.
\newblock ISBN 978-3-030-58536-5.
\newblock \doi{10.1007/978-3-030-58536-5_24}.

\bibitem[Xu et~al.(2022)Xu, Zhang, Cai, Rezatofighi, and Tao]{xu2022GMFlowLearningOpticala}
Haofei Xu, Jing Zhang, Jianfei Cai, Hamid Rezatofighi, and Dacheng Tao.
\newblock {{GMFlow}}: {{Learning Optical Flow}} via {{Global Matching}}.
\newblock In \emph{Proceedings of the {{IEEE}}/{{CVF Conference}} on {{Computer Vision}} and {{Pattern Recognition}}}, pages 8121--8130, 2022.

\bibitem[Huang et~al.(2022)Huang, Shi, Zhang, Wang, Cheung, Qin, Dai, and Li]{huang2022FlowFormerTransformerArchitecturea}
Zhaoyang Huang, Xiaoyu Shi, Chao Zhang, Qiang Wang, Ka~Chun Cheung, Hongwei Qin, Jifeng Dai, and Hongsheng Li.
\newblock {{FlowFormer}}: {{A Transformer Architecture}} for~{{Optical Flow}}.
\newblock In Shai Avidan, Gabriel Brostow, Moustapha Ciss{\'e}, Giovanni~Maria Farinella, and Tal Hassner, editors, \emph{Computer {{Vision}} -- {{ECCV}} 2022}, pages 668--685, Cham, 2022. Springer Nature Switzerland.
\newblock ISBN 978-3-031-19790-1.
\newblock \doi{10.1007/978-3-031-19790-1_40}.

\bibitem[Wang and Deng(2025)]{wang2025WAFTWarpingAloneField}
Yihan Wang and Jia Deng.
\newblock {{WAFT}}: {{Warping-Alone Field Transforms}} for {{Optical Flow}}, October 2025.

\bibitem[Wertheimer(1938)]{wertheimer1938LawsOrganizationPerceptual}
Max Wertheimer.
\newblock Laws of organization in perceptual forms.
\newblock In \emph{A Source Book of {{Gestalt}} Psychology}, pages 71--88. Kegan Paul, Trench, Trubner \& Company, London, England, 1938.
\newblock \doi{10.1037/11496-005}.

\bibitem[Yuille and Grzywacz(1988)]{yuille1988ComputationalTheoryPerception}
Alan~L. Yuille and Norberto~M. Grzywacz.
\newblock A computational theory for the perception of coherent visual motion.
\newblock \emph{Nature}, 333\penalty0 (6168):\penalty0 71--74, May 1988.
\newblock ISSN 1476-4687.
\newblock \doi{10.1038/333071a0}.

\bibitem[Liu et~al.(2021)Liu, Wu, Yu, and Lin]{liu2021EmergenceObjectness}
Runtao Liu, Zhirong Wu, Stella~X. Yu, and Stephen Lin.
\newblock The emergence of objectness: Learning zero-shot segmentation from videos.
\newblock In \emph{Advances in Neural Information Processing Systems}, volume~34, 2021.
\newblock URL \url{https://papers.neurips.cc/paper/2021/hash/6d9cb7de5e8ac30bd5e8734bc96a35c1-Abstract.html}.

\bibitem[Lian et~al.(2023)Lian, Wu, and Yu]{lian2023BootstrappingObjectness}
Long Lian, Zhirong Wu, and Stella~X. Yu.
\newblock Bootstrapping objectness from videos by relaxed common fate and visual grouping.
\newblock In \emph{Proceedings of the IEEE/CVF Conference on Computer Vision and Pattern Recognition (CVPR)}, pages 14582--14591, 2023.
\newblock \doi{10.1109/CVPR52729.2023.01401}.
\newblock URL \url{https://openaccess.thecvf.com/content/CVPR2023/html/Lian_Bootstrapping_Objectness_From_Videos_by_Relaxed_Common_Fate_and_Visual_CVPR_2023_paper.html}.

\bibitem[Dosovitskiy et~al.(2015)Dosovitskiy, Fischer, Ilg, Hausser, Hazirbas, Golkov, {van der Smagt}, Cremers, and Brox]{dosovitskiy2015FlowNetLearningOptical}
Alexey Dosovitskiy, Philipp Fischer, Eddy Ilg, Philip Hausser, Caner Hazirbas, Vladimir Golkov, Patrick {van der Smagt}, Daniel Cremers, and Thomas Brox.
\newblock {{FlowNet}}: {{Learning Optical Flow With Convolutional Networks}}.
\newblock In \emph{Proceedings of the {{IEEE International Conference}} on {{Computer Vision}}}, pages 2758--2766, 2015.

\bibitem[Mayer et~al.(2016)Mayer, Ilg, Hausser, Fischer, Cremers, Dosovitskiy, and Brox]{mayer2016LargeDatasetTrain}
Nikolaus Mayer, Eddy Ilg, Philip Hausser, Philipp Fischer, Daniel Cremers, Alexey Dosovitskiy, and Thomas Brox.
\newblock A {{Large Dataset}} to {{Train Convolutional Networks}} for {{Disparity}}, {{Optical Flow}}, and {{Scene Flow Estimation}}.
\newblock In \emph{Proceedings of the {{IEEE Conference}} on {{Computer Vision}} and {{Pattern Recognition}}}, pages 4040--4048, 2016.

\bibitem[Butler et~al.(2012)Butler, Wulff, Stanley, and Black]{butler2012NaturalisticOpenSource}
Daniel~J. Butler, Jonas Wulff, Garrett~B. Stanley, and Michael~J. Black.
\newblock A {{Naturalistic Open Source Movie}} for {{Optical Flow Evaluation}}.
\newblock In Andrew Fitzgibbon, Svetlana Lazebnik, Pietro Perona, Yoichi Sato, and Cordelia Schmid, editors, \emph{Computer {{Vision}} -- {{ECCV}} 2012}, pages 611--625, Berlin, Heidelberg, 2012. Springer.
\newblock ISBN 978-3-642-33783-3.
\newblock \doi{10.1007/978-3-642-33783-3_44}.

\bibitem[Menze and Geiger(2015)]{menze2015ObjectSceneFlow}
Moritz Menze and Andreas Geiger.
\newblock Object {{Scene Flow}} for {{Autonomous Vehicles}}.
\newblock In \emph{Proceedings of the {{IEEE Conference}} on {{Computer Vision}} and {{Pattern Recognition}}}, pages 3061--3070, 2015.

\bibitem[Sun et~al.(2018)Sun, Yang, Liu, and Kautz]{sun2018PWCNetCNNsOpticala}
Deqing Sun, Xiaodong Yang, Ming-Yu Liu, and Jan Kautz.
\newblock {{PWC-Net}}: {{CNNs}} for {{Optical Flow Using Pyramid}}, {{Warping}}, and {{Cost Volume}}.
\newblock In \emph{Proceedings of the {{IEEE Conference}} on {{Computer Vision}} and {{Pattern Recognition}}}, pages 8934--8943, 2018.

\bibitem[Dong et~al.(2023)Dong, Cao, and Fu]{dong2023RethinkingOpticalFlow}
Qiaole Dong, Chenjie Cao, and Yanwei Fu.
\newblock Rethinking {{Optical Flow}} from {{Geometric Matching Consistent Perspective}}.
\newblock In \emph{2023 {{IEEE}}/{{CVF Conference}} on {{Computer Vision}} and {{Pattern Recognition}} ({{CVPR}})}, pages 1337--1347, Vancouver, BC, Canada, June 2023. IEEE.
\newblock ISBN 979-8-3503-0129-8.
\newblock \doi{10.1109/CVPR52729.2023.00135}.

\bibitem[Zhou et~al.(2024)Zhou, He, Tan, and Yan]{zhou2024SAMFlowEliminatingAny}
Shili Zhou, Ruian He, Weimin Tan, and Bo~Yan.
\newblock {{SAMFlow}}: {{Eliminating Any Fragmentation}} in {{Optical Flow}} with {{Segment Anything Model}}.
\newblock \emph{Proceedings of the AAAI Conference on Artificial Intelligence}, 38\penalty0 (7):\penalty0 7695--7703, March 2024.
\newblock ISSN 2374-3468.
\newblock \doi{10.1609/aaai.v38i7.28603}.

\bibitem[Wang et~al.(2025)Wang, Lipson, and Deng]{wang2025SEARAFTSimpleEfficient}
Yihan Wang, Lahav Lipson, and Jia Deng.
\newblock {{SEA-RAFT}}: {{Simple}}, {{Efficient}}, {{Accurate RAFT}} for~{{Optical Flow}}.
\newblock In Ale{\v s} Leonardis, Elisa Ricci, Stefan Roth, Olga Russakovsky, Torsten Sattler, and G{\"u}l Varol, editors, \emph{Computer {{Vision}} -- {{ECCV}} 2024}, pages 36--54, Cham, 2025. Springer Nature Switzerland.
\newblock ISBN 978-3-031-72667-5.
\newblock \doi{10.1007/978-3-031-72667-5_3}.

\bibitem[Luo et~al.(2024)Luo, Li, Yang, Liu, Fan, and Liu]{luo2024FlowDiffuserAdvancingOptical}
Ao~Luo, Xin Li, Fan Yang, Jiangyu Liu, Haoqiang Fan, and Shuaicheng Liu.
\newblock {{FlowDiffuser}}: {{Advancing Optical Flow Estimation}} with {{Diffusion Models}}.
\newblock In \emph{Proceedings of the {{IEEE}}/{{CVF Conference}} on {{Computer Vision}} and {{Pattern Recognition}}}, pages 19167--19176, 2024.

\bibitem[Jiang et~al.(2021)Jiang, Campbell, Lu, Li, and Hartley]{jiang2021LearningEstimateHidden}
Shihao Jiang, Dylan Campbell, Yao Lu, Hongdong Li, and Richard Hartley.
\newblock Learning {{To Estimate Hidden Motions With Global Motion Aggregation}}.
\newblock In \emph{Proceedings of the {{IEEE}}/{{CVF International Conference}} on {{Computer Vision}}}, pages 9772--9781, 2021.

\bibitem[Morimitsu et~al.(2025)Morimitsu, Zhu, Cesar, Ji, and Yin]{morimitsu2025DPFlowAdaptiveOptical}
Henrique Morimitsu, Xiaobin Zhu, Roberto~M. Cesar, Xiangyang Ji, and Xu-Cheng Yin.
\newblock {{DPFlow}}: {{Adaptive Optical Flow Estimation}} with a {{Dual-Pyramid Framework}}.
\newblock In \emph{Proceedings of the {{IEEE}}/{{CVF Conference}} on {{Computer Vision}} and {{Pattern Recognition}}}, pages 17810--17820, 2025.

\bibitem[Jung et~al.(2023)Jung, Hui, Luo, Yang, Liu, Yoo, Ranjan, and Demandolx]{jung2023AnyFlowArbitraryScale}
Hyunyoung Jung, Zhuo Hui, Lei Luo, Haitao Yang, Feng Liu, Sungjoo Yoo, Rakesh Ranjan, and Denis Demandolx.
\newblock {{AnyFlow}}: {{Arbitrary Scale Optical Flow With Implicit Neural Representation}}.
\newblock In \emph{Proceedings of the {{IEEE}}/{{CVF Conference}} on {{Computer Vision}} and {{Pattern Recognition}}}, pages 5455--5465, 2023.

\bibitem[Achanta et~al.(2012)Achanta, Shaji, Smith, Lucchi, Fua, and S{\"u}sstrunk]{achanta2012SLICSuperpixelsCompared}
Radhakrishna Achanta, Appu Shaji, Kevin Smith, Aurelien Lucchi, Pascal Fua, and Sabine S{\"u}sstrunk.
\newblock {{SLIC Superpixels Compared}} to {{State-of-the-Art Superpixel Methods}}.
\newblock \emph{IEEE Transactions on Pattern Analysis and Machine Intelligence}, 34\penalty0 (11):\penalty0 2274--2282, November 2012.
\newblock ISSN 1939-3539.
\newblock \doi{10.1109/TPAMI.2012.120}.

\bibitem[{Van den Bergh} et~al.(2012){Van den Bergh}, Boix, Roig, {de Capitani}, and Van~Gool]{vandenbergh2012SEEDSSuperpixelsExtracted}
Michael {Van den Bergh}, Xavier Boix, Gemma Roig, Benjamin {de Capitani}, and Luc Van~Gool.
\newblock {{SEEDS}}: {{Superpixels Extracted}} via {{Energy-Driven Sampling}}.
\newblock In Andrew Fitzgibbon, Svetlana Lazebnik, Pietro Perona, Yoichi Sato, and Cordelia Schmid, editors, \emph{Computer {{Vision}} -- {{ECCV}} 2012}, pages 13--26, Berlin, Heidelberg, 2012. Springer.
\newblock ISBN 978-3-642-33786-4.
\newblock \doi{10.1007/978-3-642-33786-4_2}.

\bibitem[Jampani et~al.(2018)Jampani, Sun, Liu, Yang, and Kautz]{jampani2018SuperpixelSamplingNetworks}
Varun Jampani, Deqing Sun, Ming-Yu Liu, Ming-Hsuan Yang, and Jan Kautz.
\newblock Superpixel {{Sampling Networks}}, July 2018.

\bibitem[Yang et~al.(2020)Yang, Sun, Jin, and Zhou]{yang2020SuperpixelSegmentationFully}
Fengting Yang, Qian Sun, Hailin Jin, and Zihan Zhou.
\newblock Superpixel {{Segmentation With Fully Convolutional Networks}}.
\newblock In \emph{Proceedings of the {{IEEE}}/{{CVF Conference}} on {{Computer Vision}} and {{Pattern Recognition}}}, pages 13964--13973, 2020.

\bibitem[Sim{\'e}oni et~al.(2025)Sim{\'e}oni, Vo, Seitzer, Baldassarre, Oquab, Jose, Khalidov, Szafraniec, Yi, Ramamonjisoa, Massa, Haziza, Wehrstedt, Wang, Darcet, Moutakanni, Sentana, Roberts, Vedaldi, Tolan, Brandt, Couprie, Mairal, J{\'e}gou, Labatut, and Bojanowski]{simeoni2025DINOv3}
Oriane Sim{\'e}oni, Huy~V. Vo, Maximilian Seitzer, Federico Baldassarre, Maxime Oquab, Cijo Jose, Vasil Khalidov, Marc Szafraniec, Seungeun Yi, Micha{\"e}l Ramamonjisoa, Francisco Massa, Daniel Haziza, Luca Wehrstedt, Jianyuan Wang, Timoth{\'e}e Darcet, Th{\'e}o Moutakanni, Leonel Sentana, Claire Roberts, Andrea Vedaldi, Jamie Tolan, John Brandt, Camille Couprie, Julien Mairal, Herv{\'e} J{\'e}gou, Patrick Labatut, and Piotr Bojanowski.
\newblock {{DINOv3}}, August 2025.

\bibitem[Yang et~al.(2024)Yang, Kang, Huang, Zhao, Xu, Feng, and Zhao]{yang2024DepthAnythingV2}
Lihe Yang, Bingyi Kang, Zilong Huang, Zhen Zhao, Xiaogang Xu, Jiashi Feng, and Hengshuang Zhao.
\newblock Depth {{Anything V2}}.
\newblock \emph{Advances in Neural Information Processing Systems}, 37:\penalty0 21875--21911, December 2024.
\newblock \doi{10.52202/079017-0688}.

\bibitem[Ilg et~al.(2017)Ilg, Mayer, Saikia, Keuper, Dosovitskiy, and Brox]{ilg2017FlowNet20Evolution}
Eddy Ilg, Nikolaus Mayer, Tonmoy Saikia, Margret Keuper, Alexey Dosovitskiy, and Thomas Brox.
\newblock {{FlowNet}} 2.0: {{Evolution}} of {{Optical Flow Estimation With Deep Networks}}.
\newblock In \emph{Proceedings of the {{IEEE Conference}} on {{Computer Vision}} and {{Pattern Recognition}}}, pages 2462--2470, 2017.

\bibitem[Sun et~al.(2022)Sun, Chen, Zhu, Guo, and Li]{sun2022SKFlowLearningOpticala}
Shangkun Sun, Yuanqi Chen, Yu~Zhu, Guodong Guo, and Ge~Li.
\newblock {{SKFlow}}: {{Learning Optical Flow}} with {{Super Kernels}}.
\newblock In \emph{Advances in {{Neural Information Processing Systems}}}, 2022.

\bibitem[Zheng et~al.(2022)Zheng, Nie, Ling, Xiong, Liu, Wang, and Li]{zheng2022DIPDeepInverse}
Zihua Zheng, Ni~Nie, Zhi Ling, Pengfei Xiong, Jiangyu Liu, Hao Wang, and Jiankun Li.
\newblock {{DIP}}: {{Deep Inverse Patchmatch}} for {{High-Resolution Optical Flow}}.
\newblock In \emph{Proceedings of the {{IEEE}}/{{CVF Conference}} on {{Computer Vision}} and {{Pattern Recognition}}}, pages 8925--8934, 2022.

\bibitem[Sui et~al.(2022)Sui, Li, Geng, Wu, Xu, Liu, Goh, and Zhu]{sui2022CRAFTCrossAttentionalFlow}
Xiuchao Sui, Shaohua Li, Xue Geng, Yan Wu, Xinxing Xu, Yong Liu, Rick Goh, and Hongyuan Zhu.
\newblock {{CRAFT}}: {{Cross-Attentional Flow Transformer}} for {{Robust Optical Flow}}.
\newblock In \emph{Proceedings of the {{IEEE}}/{{CVF Conference}} on {{Computer Vision}} and {{Pattern Recognition}}}, pages 17602--17611, 2022.

\bibitem[Zhao et~al.(2022)Zhao, Zhao, Zhang, Zhou, and Metaxas]{zhao2022GlobalMatchingOverlapping}
Shiyu Zhao, Long Zhao, Zhixing Zhang, Enyu Zhou, and Dimitris Metaxas.
\newblock Global {{Matching}} with {{Overlapping Attention}} for {{Optical Flow Estimation}}.
\newblock In \emph{2022 {{IEEE}}/{{CVF Conference}} on {{Computer Vision}} and {{Pattern Recognition}} ({{CVPR}})}, pages 17571--17580, New Orleans, LA, USA, June 2022. IEEE.
\newblock ISBN 978-1-6654-6946-3.
\newblock \doi{10.1109/CVPR52688.2022.01707}.

\bibitem[Deng et~al.(2023)Deng, Luo, Huang, Ma, Liu, and Liu]{deng2023ExplicitMotionDisentangling}
Changxing Deng, Ao~Luo, Haibin Huang, Shaodan Ma, Jiangyu Liu, and Shuaicheng Liu.
\newblock Explicit {{Motion Disentangling}} for {{Efficient Optical Flow Estimation}}.
\newblock In \emph{2023 {{IEEE}}/{{CVF International Conference}} on {{Computer Vision}} ({{ICCV}})}, pages 9487--9496, Paris, France, October 2023. IEEE.
\newblock ISBN 979-8-3503-0718-4.
\newblock \doi{10.1109/ICCV51070.2023.00873}.

\bibitem[Shi et~al.(2023)Shi, Huang, Li, Zhang, Cheung, See, Qin, Dai, and Li]{shi2023FlowFormerMaskedCost}
Xiaoyu Shi, Zhaoyang Huang, Dasong Li, Manyuan Zhang, Ka~Chun Cheung, Simon See, Hongwei Qin, Jifeng Dai, and Hongsheng Li.
\newblock {{FlowFormer}}++: {{Masked Cost Volume Autoencoding}} for {{Pretraining Optical Flow Estimation}}.
\newblock In \emph{2023 {{IEEE}}/{{CVF Conference}} on {{Computer Vision}} and {{Pattern Recognition}} ({{CVPR}})}, pages 1599--1610, Vancouver, BC, Canada, June 2023. IEEE.
\newblock ISBN 979-8-3503-0129-8.
\newblock \doi{10.1109/CVPR52729.2023.00160}.

\bibitem[Morimitsu et~al.(2024)Morimitsu, Zhu, Ji, and Yin]{morimitsu2024RecurrentPartialKernel}
Henrique Morimitsu, Xiaobin Zhu, Xiangyang Ji, and Xu-Cheng Yin.
\newblock Recurrent {{Partial Kernel Network}} for {{Efficient Optical Flow Estimation}}.
\newblock \emph{Proceedings of the AAAI Conference on Artificial Intelligence}, 38\penalty0 (5):\penalty0 4278--4286, March 2024.
\newblock ISSN 2374-3468.
\newblock \doi{10.1609/aaai.v38i5.28224}.

\bibitem[Jahedi et~al.(2024)Jahedi, Luz, Rivinius, and Bruhn]{jahedi2024CCMRHighResolution}
Azin Jahedi, Maximilian Luz, Marc Rivinius, and Andr{\'e}s Bruhn.
\newblock {{CCMR}}: {{High Resolution Optical Flow Estimation}} via {{Coarse-To-Fine Context-Guided Motion Reasoning}}.
\newblock In \emph{Proceedings of the {{IEEE}}/{{CVF Winter Conference}} on {{Applications}} of {{Computer Vision}}}, pages 6899--6908, 2024.

\end{thebibliography}
\end{document}